\documentclass[conference]{IEEEtran}
\IEEEoverridecommandlockouts

\usepackage{textcomp}

\usepackage{microtype}
\usepackage{graphicx}
\usepackage{tabularx}
\usepackage{multicol}
\usepackage{booktabs} 
\usepackage{subfig}

\usepackage{amsmath}
\usepackage{amssymb}
\usepackage{mathtools}
\usepackage{amsthm}
\usepackage{amsfonts}

\theoremstyle{plain}
\newtheorem{theorem}{Theorem}[section]

\theoremstyle{definition}
\newtheorem{definition}[theorem]{Definition}

\theoremstyle{remark}

\usepackage{algpseudocode}
\usepackage{algorithm}
\usepackage{multirow}
\usepackage{array}
\usepackage{enumitem}
\usepackage{xcolor}
\usepackage{mathtools}
\usepackage{amsthm}
\usepackage{bbm}
\usepackage{wrapfig,lipsum,booktabs}
\usepackage{newfloat}
\usepackage{listings}
\usepackage{bm}
\usepackage{microtype}
\usepackage{caption}
\usepackage{url}

\newcommand{\Prob}{\mathcal{P}}

\AtBeginDocument{%
  \providecommand\BibTeX{{%
    \normalfont B\kern-0.5em{\scshape i\kern-0.25em b}\kern-0.8em\TeX}}}
    
\begin{document}

\title{Gaussian Process Tilted Nonparametric Density Estimation using Fisher Divergence Score Matching}

\author{
\IEEEauthorblockN{John Paisley}
\IEEEauthorblockA{\textit{Columbia University}}
\and
\IEEEauthorblockN{Wei Zhang}
\IEEEauthorblockA{\textit{Columbia University}}
\and
\IEEEauthorblockN{Brian Barr}
\IEEEauthorblockA{\textit{Capital One Labs}}
}

\maketitle

\begin{abstract}
We propose a nonparametric density estimator based on the Gaussian process (GP) and derive three novel closed form learning algorithms based on Fisher divergence (FD) score matching. The density estimator is formed by multiplying a base multivariate normal distribution with an exponentiated GP refinement, and so we refer to it as a GP-tilted nonparametric density. By representing the GP part of the score as a linear function using the random Fourier feature (RFF) approximation, we show that optimization can be solved in closed form for the three FD-based objectives considered. This includes the basic and noise conditional versions of the Fisher divergence, as well as an alternative to noise conditional FD models based on variational inference (VI) that we propose in this paper. For this novel learning approach, we propose an ELBO-like optimization to approximate the posterior distribution, with which we then derive a \textit{Fisher variational predictive distribution}. The RFF representation of the GP, which is functionally equivalent to a single layer neural network score model with cosine activation, provides a useful linear representation of the GP for which all expectations can be solved. The Gaussian base distribution also helps with tractability of the VI approximation and ensures that our proposed density is well-defined. We demonstrate our three learning algorithms, as well as a MAP baseline algorithm, on several low dimensional density estimation problems. The closed form nature of the learning problem removes the reliance on iterative learning algorithms, making this technique particularly well-suited to big data sets, since only sufficient statistics collected from a single pass through the data is needed.
\end{abstract}

\section{Introduction}

Density estimation is a fundamental aspect of generative modeling. With neural networks, density estimation has been significantly improved for generative models using autoregressive models \cite{larochelle2011neural,germain2015made, van2016conditional,van2016wavenet}, VAEs \cite{kingma2013auto,rezende2014stochastic,zhang2025explainable}, energy-based models \cite{song2021train}, normalizing flows \cite{rezende2015variational, dinh2016density,kingma2016improved,kingma2018glow,chen2025entropy} and GANs  \cite{goodfellow2014generative}. Parallel to these developments, score-based methods have emerged as a powerful learning approach for generative models \cite{song2019generative}. In place of the log-likelihood function, it uses the Fisher divergence between the model and data distributions to match the log gradients of the two distributions,
\begin{equation}\label{eq.div1}
    \mathbb{E}_{x\sim p_{\mathrm{data}}(x)}\big[\|\nabla_x \log p(x) - s_\theta(x)\|^2\big].
\end{equation}
The gradient of the log density approximation, $s_\theta(x)$, also called the score function \cite{liu2016kernelized}, matches those of the true data distribution. Often this $s_\theta(x)$ is a complex neural network with parameters $\theta$. Many techniques have been proposed to optimize this divergence \cite{hyvarinen2005estimation,vincent2011connection,song2020sliced} and samples are generated from $p_\theta(x)\propto\exp\{\int s_\theta(x)dx\}$ with Langevin dynamics. {To avoid an indefinite integral we will opt to write $\nabla_x s_\theta(x)$ in (\ref{eq.div1}).}

The advantages of using Fisher divergence for score matching have been demonstrated for breaking the curse of dimensionality \cite{cole2024score} and solving general inverse problems \cite{song2021solving}, such as compressed sensing MRI \cite{jalal2021robust}. An additional benefit of using score matching is that the score function $s_\theta(x)$ does not need to correspond to a textbook density and so any neural network model can be used, in effect making the density estimation nonparmetric. While therefore clearly very powerful, the common limitations of choosing and learning complex neural networks of many parameters arises. 

In this paper, we focus on lower density estimation problems for which a simpler network is sufficient. In particular, we base our score function on a proposed Gaussian process tilted density. Gaussian processes (GP) and tilted distributions have both been shown to work well for nonparametric density estimation problems \cite{hensman2014tilted,dutordoir2018gaussian,floto2023tilted}. Kernel-based exponential family ideas have also been introduced to the score matching framework, leading to iterative learning algorithms and deep representations \cite{wenliang2019learning,wang2020wasserstein,pacchiardi2022score,tsymboi2022denoising}. In contrast with these methods, our density will lead to non-iterative algorithms that calculate closed form solutions.

To achieve this, we propose a GP-tilted density estimator using the random Fourier feature approximation and learn it via Fisher divergence minimization. Our selected GP is mathematically equivalent to a single layer fully connected neural network with randomized parameters; the only parameter to be learned is a linear layer vector with dimension equal to the approximation level of the GP. We derive a noise conditional method and a novel variational predictive distribution for learning, both of which have closed form solutions as a result of our chosen score function. We evaluate our Fisher divergence-based algorithms, as well as MAP and kernel density estimation, on several low dimensional problems.

\vspace{10pt}\noindent The main features of our Tilted Gaussian Process (TGP) density approximation method are two-fold:
\begin{enumerate}[leftmargin=*]
    \item Our score function is equivalent to a single hidden layer neural network for which one parameter vector needs to be learned due to randomization of the hidden layer parameters. We present three different algorithms for this based on modifications of the Fisher divergence, all of which give closed form solutions for this vector, making the technique particularly well-suited for efficiently learning with big data through parallel computing.
    \item We propose a new approach to Fisher divergence minimization based on an approximation to the model posterior using an ELBO-like function and variational tempering. A \textit{Fisher variational predictive distribution} on the data integrates out uncertainty in the model parameter rather than by adding noise to the data, giving an alternative to noise conditional Fisher divergence methods for our model.\vspace{10pt}
\end{enumerate}
The paper is organized as follows: In Section 2, we provide a brief review of aspects of the Gaussian process and the Fisher divergence score matching methods that are relevant for our derivations. In Section 3, we present the GP-tilted density model (TGP) and derive three different, closed form solutions for learning its parameter vector. We then demonstrate the model on several low dimensional data sets in Section 4.

\section{Background}
In this section, we briefly review background material on Gaussian processes and Fisher divergence minimization useful for our GP-tilted density estimation model and algorithms.

\subsection{Gaussian Processes}\label{sec.gpback}
The Gaussian process (GP) is a core nonparametric function approximation technique in probabilistic modeling and machine learning \cite{williams2006gaussian}. Its formal definition is as follows:

\begin{definition}[Gaussian Process]
Let $k(x,x')$ be a kernel function between two vectors $x,x'\in\mathbb{R}^d$. Then the function $f : \mathbb{R}^d \rightarrow \mathbb{R}$ is a zero-mean Gaussian process if, for any $N$ vectors $x_1,\dots,x_N$, the corresponding function evaluations $f(x_1),\dots,f(x_N)$ are distributed as $\mathcal{N}(0,K_N)$, where $K_N(i,j) = k(x_i,x_j)$. We write $f(x) \sim \mathcal{GP}(0,k(x,x'))$.
\end{definition}

There are a large variety of kernel functions $k$ that can be used with the GP, and extensions can be made to other spaces and non-zero mean functions. The most popular kernel for machine learning applications is the Gaussian kernel of the form $k(x,x') = \lambda\exp(-\frac{1}{2\gamma^2}\|x-x'\|^2)$, which measures correlations based on proximity in the input space. For the Gaussian kernel, the following random Fourier feature (RFF) definition provides an asymptotically exact way of representing the underlying linear model of the GP in its reproducing kernel Hilbert space  \cite{rahimi}. The equivalent structure of this representation to a single layer of a neural network has recently motivated other applications of this idea to traditional machine learning problems \cite{paisley2022bayesian,zhang2024a,zhang2024b}. 

\begin{definition}[Random Fourier Features] \label{def.rff}
Let $S$ to be a positive integer, $\gamma > 0$, and $x\in\mathbb{R}^d$. For $s=1,\dots,S$, generate $z_s\sim \mathcal{N}(0,I)$ and $c_s\sim \mathrm{Unif}(0,2\pi)$. Define a random Fourier feature mapping $\phi$ of the vector $x$ as
$$\phi(x) = \sqrt{\frac{2}{S}}\left[\cos{\left(\frac{z_1^\top x}{\gamma} +c_1\right)},\dots,\cos{\left(\frac{z_S^\top x}{\gamma} + c_S\right)}\right]^\top .$$
Then, for any $x,x'\in\mathbb{R}^d$, $\phi(x)^\top\phi(x') \rightarrow \exp(-\frac{1}{2\gamma^2}\|x-x'\|^2)$ as  $S\rightarrow\infty$. Thus $\phi$ approximates the RKHS of the GP.
\end{definition}

It follows from the marginal representation of the Gaussian process \cite{williams2006gaussian} that, if $\theta \sim \mathcal{N}(0,\lambda^{-1} I)$, then the random function $f_\theta(x) = \theta^\top \phi(x)$ is distributed as $\mathcal{GP}(0,k(x,x'))$ with 
$\textstyle k(x,x') = \lambda^{-1}\phi(x)^\top\phi(x')$. Since this dot product converges to $\exp(-\frac{1}{2\gamma^2}\|x-x'\|^2)$ as $S\rightarrow\infty$, the approximation holds.

\subsection{Fisher Divergence for Parameter Learning}
Let $\Prob$ be the true unknown distribution of data from which we have i.i.d.\ samples, $x\sim\Prob$. Our goal is to approximate the density $p(x)$ corresponding to $\Prob$ with another density $q_\theta(x)$. This is achieved by tuning parameters $\theta$ such that the two densities agree according to some measure of similarity. Maximum likelihood is one standard approach based on the KL-divergence,
$$\theta_{\textrm{ML}} = \arg\min_{\theta} \mathbb{E}_p\Big[\ln\frac{p(x)}{q_{\theta}(x)}\Big] \approx \arg\max_{\theta} \sum\nolimits_{i} \ln q_{\theta}(x_i).$$
Here, the data provides a Monte Carlo approximation.

A second method that has recently been effective for learning deep generative models is the Fisher divergence, which optimizes $\theta$ as follows,
\begin{equation}
 \theta_{\textrm{FD}} = \arg\min_{\theta} \tfrac{1}{2} \mathbb{E}_p\left[\|\nabla_x \ln p(x) - \nabla_x \ln q_{\theta}(x)\|^2\right],
\end{equation}
where a score function, typically a neural network and written here as $\nabla_x\ln q_{\theta}(x)$, is designed to match the gradients of the log data distribution with emphasis on more probable regions.
Through a derivation involving integration by parts, this is equivalent to an optimization problem of the form
\begin{eqnarray}\label{eq.FDobj}
   \quad \theta_{\textrm{FD}} \hspace{-5pt}&\approx&\hspace{-5pt} \arg\min_{\theta} ~ \tfrac{1}{2}\sum\nolimits_{i=1}^N \big\|\nabla_x\ln q_{\theta}(x_i)\big\|^2 \qquad\qquad\\
    &&\qquad\quad\, + \sum\nolimits_{i=1}^N \mathrm{trace}\big( \nabla_{xx}^2 \ln q_{\theta}(x_i)\big).\nonumber
\end{eqnarray}
A Monte Carlo approximated is again made using i.i.d.\ samples $x\sim\mathcal{P}$. As mentioned, the Fisher score focuses on creating agreement between the derivatives of $q_{\theta}$ and the true distribution $p$ in high density regions. Equation \ref{eq.FDobj} shows that this is done with  $q_{\theta}$ by placing a mode around each $x_i$. 

For the GP-tilted density estimator discussed next, the Fisher divergence has the advantage of producing a closed form solution for $\theta$, in contrast to maximum likelihood (see Section \ref{sec.basic}). As observed with deep generative models, the same practical issue of accurately learning the low density regions arises, which we address in two ways: In Section \ref{sec.noise} we adopt the frequently used noise conditional representation, which still has a closed form solution for our model, and in Section \ref{sec.vi} we propose a variational approximation to the predictive distribution that also has a closed form solution.

\section{GP-Tilted Density Estimation}\label{sec.gptde}

\begin{figure}[t]
    \includegraphics[width=.95\columnwidth]{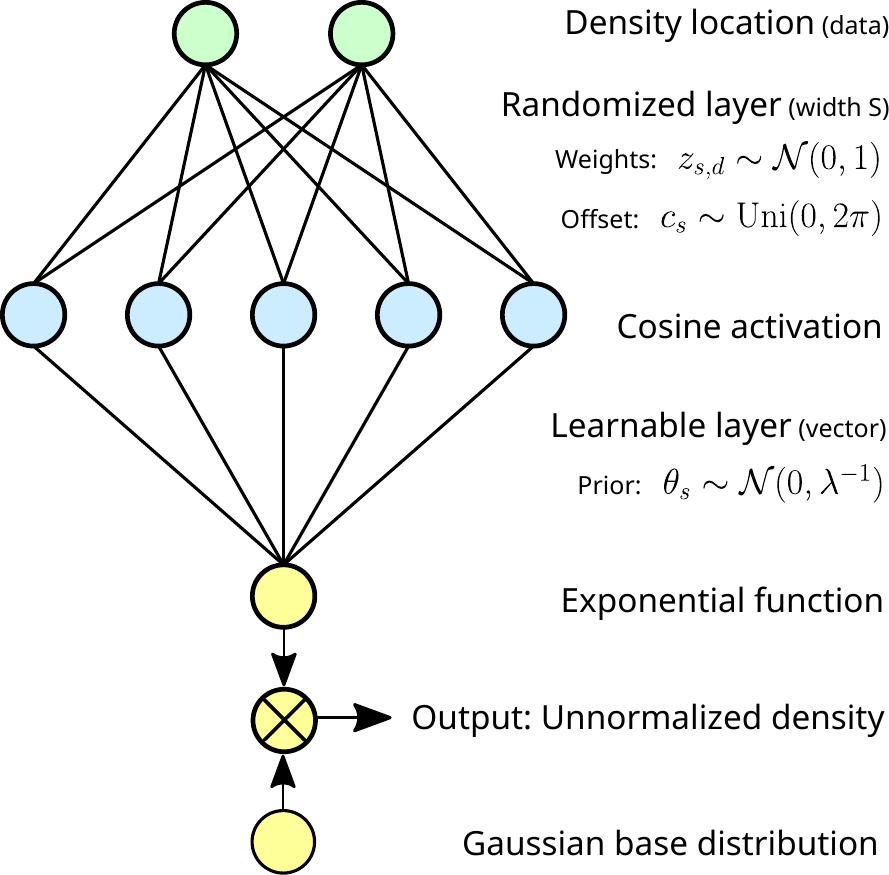}\vspace{10pt}
    \caption{A simplified flowchart of the GP-tilted density. The input to the exponential function is the Gaussian process approximated by random Fourier features, which turns the GP into a single layer neural network. The only learnable parameter is the vector $\theta \in \mathbb{R}^S$. (We set $S=1000$ in our experiments.) A broad, predefined Gaussian based distribution determines the region of interest for the density to ensure integrability. In Algorithms 2-4, we present novel closed form solutions for learning $\theta$ and constructing predictive distributions based on the Fisher divergence score matching objective function.}
    \label{fig:rffdens}
\end{figure}

Let $f(x) \sim \mathcal{GP}(0,k(x,x'))$, where $k$ is a Gaussian kernel. We define the GP-tilted (TGP) density of $x \in \mathbb{R}^d$ to be
\begin{equation}
    q(x) = \frac{\exp\lbrace{f(x)\rbrace} \mathcal{N}(x|\mu,\Sigma)}{\int \exp\lbrace{f(x)\rbrace} \mathcal{N}(x|\mu,\Sigma)\, dx}.
\end{equation}
As follows from Gaussian process theory and basic analysis, $q(x)$ is a continuous density \cite{williams2006gaussian}. We use a multivariate Gaussian base measure to ensure integrability of the numerator in $\mathbb{R}^d$. The mean $\mu$ and covariance $\Sigma$ broadly define the region of interest over the data, while the exponentiated GP provides a nonparametric refinement. 
Using the RFF representation of $f(x)$ in Definition \ref{def.rff}, this density can be approximated by
\begin{eqnarray}
q_{\theta}(x) &\propto& \exp\lbrace{\theta^\top \phi(x)\rbrace} \mathcal{N}(x|\mu,\Sigma),\nonumber\\
\theta &\sim& \mathcal{N}(0,\lambda^{-1}I),\label{eq.density}
\end{eqnarray}
where $\phi$ is the $S$-dimensional RFF mapping. MAP inference for this model is fairly straightforward, which we present in Algorithm \ref{alg:MLinference} for reference. Functionally, the score $\ln q_\theta(x)$ has the form of a single layer neural network using RFF parameters. In Figure \ref{fig:rffdens} we show a simplified flowchart of the TGP density estimator, connecting it with a single layer neural network structure. We next show that this approximation also allows for easy inference using Fisher divergence minimization.

\subsection{Fisher Divergence Score Matching}\label{sec.basic}
We recall that minimizing the Fisher divergence between the GP-tilted density and the true data density using i.i.d.\ samples entails working with the function
\begin{equation}\label{eq.objective}
\mathcal{L}= \tfrac{1}{2}\sum\nolimits_i \big\|\nabla_x\ln q_{\theta}(x_i)\big\|^2 + \sum\nolimits_i \mathrm{tr}\big( \nabla_{xx}^2 \ln q_{\theta}(x_i)\big).
\end{equation}
(We account for the prior on $\theta$ later.) The gradient and Hessian of $\ln q_{\theta}(x)$ for the TGP density are
\begin{eqnarray}\label{eq.nabx}
    \nabla_x \ln q_{\theta}(x) \hspace{-5pt}& = &\hspace{-5pt}  \sum\nolimits_s \tfrac{1}{\gamma}\theta_s \phi'_s(x) z_s  - \Sigma^{-1}(x-\mu),\\\label{eq.nabxx}
    \nabla_{xx}^2 \ln q_{\theta}(x) \hspace{-5pt}& = &\hspace{-5pt}  \sum\nolimits_s \tfrac{1}{\gamma^2}\theta_s  \phi''_s(x) z_s z_s^\top - \Sigma^{-1},
\end{eqnarray}
where for $s=1,\dots,S$ we have defined 
\begin{eqnarray}\label{eq.phiprime}
    \phi'_s(x) \hspace{-3pt}&=&\hspace{-3pt} -\sqrt{2/S}\sin(z_s^\top x/\gamma + c_s),\nonumber\\ 
    \phi''_s(x) \hspace{-3pt}&=&\hspace{-3pt} -\sqrt{2/S}\cos(z_s^\top x/\gamma + c_s).
\end{eqnarray}
After plugging Eqs.\ \ref{eq.nabx} \& \ref{eq.nabxx} into Equation \ref{eq.objective}, the Fisher divergence is observed to be a quadratic-plus-linear function of $\theta$, which has an analytic optimal solution. Define the matrix $Z = [z_1,\dots,z_S]^\top$ and for any square symmetric matrix $M$, let $\|ZM\|^2 \equiv (\|Mz_1\|^2,\dots,\|Mz_S\|^2)^\top$ --- in this subsection, $M = I$.  Optimizing $\theta$ involves working with the gradient
\begin{eqnarray}\label{eq.nablabasic}
    \nabla_{\theta}\mathcal{L} &=& \Big(\sum\nolimits_{i=1}^N \tfrac{1}{\gamma^2} ZZ^\top \odot \phi'(x_i)\phi'(x_i)^\top\Big)\theta \nonumber\\\nonumber 
    && - \sum\nolimits_{i=1}^N \tfrac{1}{\gamma}\phi'(x_i)\odot Z\Sigma^{-1}(x_i-\mu)\\
    && + \sum\nolimits_{i=1}^N\tfrac{1}{\gamma^2}\phi''(x_i) \odot \|Z\|^2,
\end{eqnarray}
where $\odot$ indicates elementwise multiplication.

\begin{algorithm}[t]
\caption{MAP inference for GP tilted density estimation}\label{alg:MLinference}
\begin{algorithmic}[1]
\Require Data $x \in\mathbb{R}^D$, $\gamma > 0$, $\lambda >0$, $S_1, S_2 \in \mathbb{Z}_+$, base parameters $\mu\in\mathbb{R}^D$, $\Sigma \in \mathbb{S}_{++}^{d}$, step size $\rho$
\State Sample $z_s \sim \mathcal{N}(0,I_d)$, $c_s \sim \mathrm{Unif}(0,2\pi)$, $s=1:S_1$
\State Set $\psi_1 = \sum_i \phi(x_i)$, sample $\zeta_j \sim \mathcal{N}(\mu,\Sigma)$, $j=1:S_2$
\For{iteration $t = 1,\dots,T$}
\State \hspace{-1pt}Set~ $\theta_{t+1} = (1-\rho\lambda)\theta_t + \rho (\psi_1 - N\psi_2)$,\\
$\begin{array}{rcl}
\textstyle \psi_2 &\hspace{-5pt}=&\hspace{-5pt} \sum_j \phi(\zeta_j) w_t(j), \\
\textstyle\qquad w_t(j) &\hspace{-5pt}=&\hspace{-5pt} \exp\lbrace\theta_t^\top \phi(\zeta_j)\rbrace \big/ \sum_{j'}\exp\lbrace\theta_t^\top \phi(\zeta_{j'})\rbrace
\end{array}$
\EndFor\\
\Return $\theta$, $Z$, $c$, $\gamma$, $\mu$, $\Sigma$
\end{algorithmic}
\end{algorithm}

We also recall that the Gaussian prior of the linearized GP is $\theta \sim \mathcal{N}(0,\lambda^{-1} I)$. This introduces an $\ell_2$ regularization term, $\theta_{\mathrm{FD}} ~=~ \arg\min_{\theta} ~ \mathcal{L}(\theta) + \frac{\lambda}{2}\theta^\top \theta .$
Finally, using the fact that $\phi'' = -\phi$, the value of $\theta$ that minimizes this function is 
\begin{eqnarray}
    \theta_{\mathrm{FD}} \hspace{-5pt}&=&\hspace{-5pt} \left(\lambda\gamma^2 I + ZZ^\top \odot \sum\nolimits_i \phi'(x_i) \phi'(x_i)^\top\right)^{-1} \times\\
    \hspace{-5pt}&&\hspace{-5pt}\left(\sum\nolimits_i \gamma\phi'(x_i)\odot Z\Sigma^{-1}(x_i-\mu) + \phi(x_i)\odot\|Z\|^2\right).\nonumber
\end{eqnarray}
This involves inverting an $S\times S$ matrix, where $S$ is the number of random Fourier features used to approximate the Gaussian process. When large values of $S$ or other numerical issues present a computational challenge, conjugate gradients can be used to quickly find $\theta_{\mathrm{FD}}$ instead. We summarize parameter estimation for the basic GP-tilted density using Fisher divergence in Algorithm \ref{alg:inference}.

\begin{algorithm}[t]
\caption{Fisher divergence for TGP learning}\label{alg:inference}
\begin{algorithmic}[1]
\Require Data $x \in\mathbb{R}^D$, kernel width $\gamma > 0$, param $\lambda >0$, $S \in \mathbb{Z}_+$, base parameters $\mu\in\mathbb{R}^D$, $\Sigma \in \mathbb{S}_{++}^{d}$
\State Sample $z_s \sim \mathcal{N}(0,I_d)$, $c_s \sim \mathrm{Unif}(0,2\pi)$, $s=1:S$
\State Def $Z = [z_1,\dots,z_S]^\top$, $\|Z\|^2 = (\|z_1\|^2,\dots,\|z_S\|^2)^\top$
\State Construct each $\phi'(x_i)$ and  $\phi(x_i) = -\phi''(x_i)$ (Eq. \ref{eq.phiprime})
\State Using all data, calculate
$$
\begin{array}{ll}
     \Phi'&\hspace{-5pt}=~\textstyle\sum_i \phi'(x_i) \phi'(x_i)^\top\\
     \psi'&\hspace{-5pt}=~\textstyle\sum_i \phi'(x_i)\odot Z\Sigma^{-1}(x_i-\mu)\\
     \psi &\hspace{-5pt}=~\textstyle\sum_i \phi(x_i)
\end{array}
$$
\State Solve either directly or, e.g., with conjugate gradients,
$$\theta_{\mathrm{FD}} = \left(\lambda\gamma^2 I + ZZ^\top \odot \Phi'\right)^{-1}\left(\gamma\psi' + \|Z\|^2\odot\psi\right).$$
\Return $\theta_{\mathrm{FD}}$, $Z$, $c$, $\gamma$, $\mu$, $\Sigma$
\end{algorithmic}
\end{algorithm}

\subsubsection{Learning $\mu$ and $\Sigma$ of base measure}
The base Gaussian measure of $q_\theta(x)$ in Equation \ref{eq.density} is important because it defines the region of focus for the Gaussian process. When set too small, the base resists the GP's ability to increase the density near data outside this region. When it is too large, it requires the GP to learn to suppress regions of no data, which is challenging because the GP $\theta^\top\phi(x)$ reverts to something like an evaluation of the random prior. This can randomly produce large positive values away from the data that dominate the overall density $q_\theta(x)$. We discuss two modifications to the Fisher divergence to address this issue in Section \ref{sec.noise} and Section \ref{sec.vi}.

For the basic Fisher divergence, one option is to set $\mu$ and $\Sigma$ to the empirical mean and scaled covariance of the data, where the scaling is done using cross-validation. We also note an option of performing coordinate descent over $\theta$, according to Equation \ref{eq.objective}, as well as $\mu$ and $\Sigma$ as follows: For $\mu = \arg\min_{\mu} \mathcal{L}$, the optimal value given $\theta$ and $\Sigma$ is
\begin{equation}\label{eq.mu_update}
\textstyle\mu = \frac{1}{N}\big[\sum_{i=1}^N x_i\big] -  \frac{1}{\gamma}\Sigma Z^\top (\theta\odot\frac{1}{N}\sum_{i=1}^N\phi'(x_i)).
\end{equation}
Differentiating to find $\Sigma = \arg\min_{\Sigma} \mathcal{L}$, we obtain the Lyapunov equation $\overline{\Sigma} \Sigma^{-1} + \Sigma^{-1} \overline{\Sigma}=Q$, where 
\begin{align}
\overline{\Sigma} =&~\textstyle \frac{1}{N}\sum_{i=1}^N (x_i-\mu)(x_i-\mu)^\top,\\
\Upsilon =&~\textstyle \frac{1}{N}\sum_{i=1}^N(\theta\odot\phi'(x_i))(x_i-\mu)^\top,\\
Q =&~ 2I + (Z^\top\Upsilon + \Upsilon^\top Z)/\gamma .
\end{align}
This also has a closed form solution using the Kronecker product as follows,
\begin{equation}\label{eq.sig_update}
    \mathrm{vec}(\Sigma^{-1}) = (\overline{\Sigma} \otimes I + I \otimes \overline{\Sigma})^{-1}\mathrm{vec}(Q).
\end{equation}
For data $x\in\mathbb{R}^d$, this matrix inverse is $\mathcal{O}(d^6)$. This is not prohibitive for the low dimensional problems we consider. As the dimensionality increases, other algorithms can be used to solve for $\Sigma$. When optimizing Equation \ref{eq.objective}, the base parameters can be learned by iterating between updates to $\theta$ (Algorithm \ref{alg:inference}) and $\mu$ and $\Sigma$ (Equations \ref{eq.mu_update} \& \ref{eq.sig_update}). Because the algorithm is now iterative, it takes significantly more time, which is a major drawback. We set $\mu$ and $\Sigma$ as previously described.

\subsection{Noise Conditional Score Matching}\label{sec.noise}
It has been observed that the approach of Section \ref{sec.basic} can have difficulty accurately learning $q_\theta$ in regions where little or no data exists. This can lead to inflated density values in low data regions as a learning artifact, especially for GP-based scores, since reversion to the model prior in these regions produce random functions with no preference between high or low density. To address this learning issue, noise conditional score matching adds Gaussian noise of arbitrary variance to the data and learns a noise-dependent score model. A single distribution is learned to approximate the data density in all noise regimes, which provides greater information to the model for learning low density regions.

The noise-added model modifies the data distribution as
$$ p(y|\sigma) = \int p(y|x,\sigma)p(x)dx \approx \frac{1}{N}\sum_{i=1}^N \mathcal{N}(y|x_i,\sigma^2 I).$$
Below, we will equivalently write a sample $y\sim p(y|\sigma)$ as $y=x+\xi$ where $x \sim \sum_{i=1}^N \frac{1}{N}\delta_{x_i}$ and $\xi \sim \mathcal{N}(0,\sigma^2I)$. 

Next, we need to modify our TGP density to account for the noise level $\sigma$. In other score models this is simply done by including $\sigma$ as an input to the neural network. The Gaussian process permits a more natural probabilistic adaptation,
\begin{eqnarray}
q_{\theta}(y|\sigma) &\propto& \exp\lbrace{\theta^\top \phi_\sigma(y)\rbrace} \mathcal{N}(y|\mu,\Sigma_\sigma),\\
\phi_{\sigma}(y) \hspace{-2pt}&=&\hspace{-2pt} \sqrt{2/S}\cos(Zy/\gamma_{\sigma} +c),\label{eq.phisig}
\end{eqnarray}
\begin{equation}
\gamma_{\sigma} \coloneqq \sqrt{\gamma^2 + \sigma^2}, \quad\Sigma_\sigma = \Sigma + \sigma^2 I.
\end{equation}
We observe that the noise added to $x$ is naturally incorporated in $q$ in two places: 1) By expanding the variance of the base Gaussian measure to expand the data region of interest, and 2) by increasing the kernel width to learn a multiscale GP. Using one $\theta$ learned for all $\sigma\geq0$, we can evaluate $q_\theta$ at arbitrary resolution, including the original $q_\theta(x)$ when $\sigma = 0$.

With this modification, the noise conditional Fisher divergence using parameter $\sigma$ is
\begin{equation}\label{eq.noiseobj}
\hspace{-12pt}
\begin{array}{rl}
\mathcal{L}_{\sigma} =& \displaystyle\tfrac{1}{2}\sum\nolimits_{i=1}^N \mathbb{E}_{p(\xi|\sigma)}\big[\|\nabla_x\ln q_{\theta}(x_i+\xi)\|^2\big]\\
&\displaystyle+ \sum\nolimits_{i=1}^N \mathbb{E}_{p(\xi|\sigma)}\big[\mathrm{trace}\big(\nabla_{xx}^2 \ln q_{\theta}(x_i+\xi)\big)\big].
\end{array}\hspace{-10pt}
\end{equation}
The form of the derivatives were given in Equations \ref{eq.nabx} and \ref{eq.nabxx}, except the new $\sigma$-dependent $\phi_\sigma=-\phi_\sigma''$ of Equation \ref{eq.phisig} is used instead, along with its derivative
\begin{equation}
    \phi'_{\sigma}(y)= -\sqrt{2/S}\sin(Zy/\gamma_{\sigma} +c) \label{eq.phisigprime}.
\end{equation}
We next show that these expectations can be solved analytically and a closed form solution exists for solving the total objective $\theta_{\mathrm{FD}\sigma} = \arg\min_\theta\mathbb{E}_{p(\sigma)}[\mathcal{L}_\sigma] + \frac{\lambda}{2}\theta^\top\theta$ for some preset discretized distribution $\sigma\sim p(\sigma)$ defined later.

\subsubsection{Derivation of expectations}
From Equation \ref{eq.noiseobj} four unique expectations arise. For the expectation of $\cos({z^\top y}/{\gamma_\sigma} + c)$, where $y = x+\xi$ and $\xi \sim \mathcal{N}(0,\sigma^2I)$, we represent the input to the cosine as follows,
$${z^\top (x+\xi)}/{\gamma_\sigma} + c ~~\stackrel{d}{=}~~ \epsilon + \chi_{\sigma},$$
$$\chi_{\sigma} \coloneqq {z^\top x}/{\gamma_\sigma}+ c,\quad \epsilon \sim \mathcal{N}\left(0,{\tfrac{\sigma^2}{\gamma^2 + \sigma^2}}\|z\|^2\right).$$
The MGF of a univariate Gaussian can then be used to derive the following expectation,
\begin{eqnarray}
\mathbb{E}[\cos(\epsilon+\chi_{\sigma})] \hspace{-5pt}&=&\hspace{-5pt} \mathrm{Re}\left(\mathrm{e}^{j \chi_{\sigma}}\mathbb{E}_{p(\epsilon|\sigma)}\left[\mathrm{e}^{j\epsilon}\right]\right)\label{eq.Ecos} \\
\hspace{-5pt}&=&\hspace{-5pt} \exp\left(-\frac{\|z\|^2}{2}\frac{\sigma^2}{\gamma^2+\sigma^2}\right)\cos(\chi_{\sigma}),\nonumber
\end{eqnarray}
where $j = \sqrt{-1}$. Similarly, $\mathbb{E}[\sin(\epsilon+\chi_{\sigma})]$ is solved using the imaginary component, which replaces $\cos$ with $\sin$ above.

A third expectation arises of the form $\mathbb{E}[\phi_{\sigma s}'\phi_{\sigma s'}']$ We use the identity
$\sin(\alpha)\sin(\beta) = \frac{1}{2}(\cos(\alpha-\beta) - \cos(\alpha+\beta))$, where $\alpha = {z_s^\top\xi}/{\gamma_\sigma} + \chi_{\sigma s}$ and $\beta = {z_{s'}^\top\xi}/{\gamma_\sigma} + \chi_{\sigma s'}$ here, and again let $\chi_{\sigma s} \coloneqq {z_s^\top x}/{\gamma_\sigma}+ c_s$.
The expectation of these cosines follows a similar derivation as above and results in\\

$\begin{array}{c}
 \mathbb{E}[\phi'_{\sigma s}(y)\phi'_{\sigma s'}(y)] =~~~\qquad\qquad\qquad\qquad\qquad\qquad\qquad\qquad \\
 \qquad\qquad\quad~~\frac{1}{S}\exp\Big(-\frac{\sigma^2\|z_s-z_{s'}\|^2}{2(\gamma^2+\sigma^2)}\Big)\cos(\chi_{\sigma s}-\chi_{\sigma s'})\\
  \qquad\quad~~~~~~ - \frac{1}{S}\exp\Big(-\frac{\sigma^2\|z_s+z_{s'}\|^2}{2(\gamma^2+\sigma^2)}\Big)\cos(\chi_{\sigma s}+\chi_{\sigma s'})\vspace{6pt}
\end{array}$
for all $s,s'\in\{1,\dots,S\}$, including $s=s'$. Using the identity $\cos(\alpha\pm\beta) = \cos\alpha\cos\beta\mp\sin\alpha\sin\beta$, this can be simplified for faster computation as shown later.

\setlength{\tabcolsep}{4pt}
\renewcommand{\arraystretch}{1.5}{
\begin{table}\normalsize
\caption{List of Expectations (see Eqs. \ref{eq.phisig}, \ref{eq.phisigprime}, \ref{eq.del1}--\ref{eq.del3})\vspace{-5pt}}\label{tab.equations}
\centering
\begin{tabular}{rcl}
\hline\hline
     $\mathbb{E}[\phi_{\sigma}(y)]$ &$=$& $\delta_\sigma\odot\phi_\sigma(x)$\\
     $\mathbb{E}[\phi'_{\sigma}(y)]$&$=$&$\delta_\sigma \odot \phi'_\sigma(x)$\\
     $\mathbb{E}[\phi'_{\sigma}(y)\phi'_{\sigma}(y)^\top]$&$=$&$\textstyle\frac{1}{2}(\Delta^-_\sigma-\Delta^+_\sigma)\odot\phi_\sigma(x)\phi_\sigma(x)^\top $\\
     &$+$&$\textstyle\frac{1}{2}(\Delta^-_\sigma+\Delta^+_\sigma)\odot\phi'_\sigma(x)\phi'_\sigma(x)^\top $ \\
    $\mathbb{E}[(Z\Sigma_\sigma^{-1} \xi)\odot\phi'_{\sigma}(y)]$&$=$&$\textstyle -\frac{\sigma^2}{\gamma_\sigma} \delta_\sigma \odot \|Z\Sigma_\sigma^{-\frac{1}{2}}\|^2\odot\phi_\sigma(x)$\vspace{3pt}\\
\hline\hline
\end{tabular}
\end{table}
}

A final expectation occurs in the cross term of the square of the gradient in Equation \ref{eq.nabx}.
This expectation takes the form $\mathbb{E}_{p(\xi|\sigma)}[\xi^\top \Sigma_\sigma^{-1}z \sin (\xi^\top z/\gamma_\sigma + \chi_{\sigma})]$.
We solve this by writing vector products in sum form over the $d$ dimensions of $\xi$ and solving separate expectations then adding. Using the shorthand $g(\xi)={\xi^\top z}/{\gamma_\sigma} + \chi_{\sigma}$, and recalling that $\chi_{\sigma}$ is also a function of $z$, we solve by using the equality
$$\mathbb{E}\left[\xi_l \sin g(\xi)\right] = -x_l\mathbb{E}\sin g(\xi) - \gamma_\sigma({\partial}/{\partial z_l})\mathbb{E}\cos g(\xi).$$
Using the expectations for $\sin$ and $\cos$ from Equation \ref{eq.Ecos} and making some cancellations, the total expectation is
\begin{eqnarray}
&&\hspace{-23pt}\sum\nolimits_{l=1}^d(\Sigma_\sigma^{-1}z)_l\mathbb{E}\left[\xi_l \sin (\xi^\top z/\gamma_\sigma + \chi_{\sigma})\right] =\\
&& \frac{\sigma^2}{\gamma_\sigma}(z^\top\Sigma_\sigma^{-1}z)\exp\Big(-\frac{\|z\|^2}{2}\frac{\sigma^2}{\gamma^2 + \sigma^2}\Big)\cos\Big(\frac{z^\top x}{\gamma_\sigma}+c\Big).\nonumber
\end{eqnarray}
We summarize all expectations in vector form in Table \ref{tab.equations}, where we use the following definitions:
\begin{eqnarray}\label{eq.del1}
    \delta_\sigma(s) \hspace{-5pt}&=&\hspace{-5pt} \exp(- \textstyle\frac{1}{2}(\sigma/\gamma_\sigma)^2\|z_s\|^2),\\ \label{eq.del2}
    \Delta^+_\sigma(s,s')\hspace{-5pt}&=&\hspace{-5pt}\exp(-\textstyle\frac{1}{2}(\sigma/\gamma_\sigma)^2\|z_s + z_{s'}\|^2),\\\label{eq.del3}
    \Delta^-_\sigma(s,s')\hspace{-5pt}&=&\hspace{-5pt}\exp(-\textstyle\frac{1}{2}(\sigma/\gamma_\sigma)^2\|z_s - z_{s'}\|^2),
\end{eqnarray}
where $\delta_\sigma$ is an $S$-dimensional vector and $\Delta_\sigma^+$ and $\Delta_\sigma^-$ are $S\times S$ matrices. Again, we use definitions described previously: $\gamma_\sigma = \sqrt{\gamma^2+\sigma^2}$, $\|ZM\|^2 \equiv (\|Mz_1\|^2,\dots,\|Mz_S\|^2)^\top$ and $\Sigma_\sigma = \Sigma + \sigma^2 I$. Also, as defined in Equation \ref{eq.phisig}, the RFF vector $\phi_\sigma$ indicates that the kernel width $\gamma_\sigma$ is used, while $(z,c)$ are shared for all values of $\sigma$.

\begin{algorithm}[t!]
\caption{Noise conditional Fisher TGP estimation}\label{alg:inference2}
\begin{algorithmic}[1]
\Require
     Data $\{x_1,\dots,x_N\}, x \in\mathbb{R}^D$.
\Statex Model parameters $\gamma > 0$, $\lambda >0$, $\mu\in\mathbb{R}^D$, $\Sigma \in \mathbb{S}_{++}^{d}$.
\Statex Approximation parameters $\sigma_{\mathrm{max}} > 0$, $\{H,S\} \in \mathbb{Z}_+$
\State Sample  $z_s \sim \mathcal{N}(0,I_d)$, $c_s \sim \mathrm{Unif}(0,2\pi)$, $s=1:S$
\State Define $Z = [z_1,\dots,z_S]^\top$ and $c = (c_1,\dots,c_S)^\top$
\State Define $\|ZM\|^2 = (\|Mz_1\|^2,\dots,\|Mz_S\|^2)^\top$, $M\in \mathbb{S}_{++}^{d}$
\State Define $\gamma_\sigma = \sqrt{\gamma^2 + \sigma^2}$ and $\Sigma_\sigma = \Sigma + \sigma^2I$ for values of $\sigma$ in the set $\Omega_\sigma = \left\{\frac{h-1}{H}\sigma_{\mathrm{max}}: h= 1,\dots, H\right\}$
\State For data $\{x_i\}_{i=1:N}$ and each $\sigma \in \Omega_\sigma$, calculate RFF vectors $\phi_\sigma(x_i)$ and $\phi'_\sigma(x_i)$ as in Eqs \ref{eq.phisig} \& \ref{eq.phisigprime}, and
\begin{eqnarray}
\Phi'_\sigma &=& \textstyle\sum_i \phi'_\sigma(x_i)\phi'_\sigma(x_i)^\top\nonumber\\
\Phi_\sigma &=& \textstyle\sum_i \phi_\sigma(x_i)\phi_\sigma(x_i)^\top\nonumber\\
\psi'_{\sigma} &=& \textstyle\sum_i \phi'_\sigma(x_i)\odot Z\Sigma_\sigma^{-1}(x_i-\mu)\nonumber\\
\psi_\sigma &=& \textstyle\sum_i \phi_\sigma(x_i)\nonumber
\end{eqnarray}
\State Using definitions of $\delta_\sigma$, $\Delta_\sigma^-$ and $\Delta_\sigma^+$ in Eqs \ref{eq.del1}--\ref{eq.del3}, calculate the matrix $A$ and vector $b$ as follows
\begin{eqnarray}
    A & = &\textstyle \sum_{\sigma\in\Omega_\sigma}  \frac{1}{2\gamma^2_\sigma}(\Delta^-_\sigma +\Delta^+_\sigma) \odot \Phi'_\sigma \nonumber\\
    &+&\textstyle  \sum_{\sigma\in\Omega_\sigma} \frac{1}{2\gamma^2_\sigma}(\Delta^-_\sigma -\Delta^+_\sigma)\odot\Phi_\sigma \nonumber\\
    b & = &\textstyle \sum_{\sigma\in\Omega_\sigma} \frac{1}{\gamma_\sigma}\delta_\sigma \odot \psi'_{\sigma} \nonumber\\
    &+&\textstyle \sum_{\sigma\in\Omega_\sigma} \frac{1}{\gamma^2_\sigma}\|Z(I-\sigma^2\Sigma_\sigma^{-1})^{\frac{1}{2}}\|^2 \odot\delta_\sigma \odot \psi_\sigma \nonumber
\end{eqnarray}
\State Solve directly or, e.g., with conjugate gradients,
$$\theta_{\mathrm{FD}\sigma} = (\lambda H I + ZZ^\top \odot A)^{-1} b$$
\Return $\theta_{\mathrm{FD}\sigma}$, $Z$, $c$, $\gamma$, $\mu$, $\Sigma$
\end{algorithmic}
\end{algorithm}

\subsubsection{Algorithm summary and discussion}
We can now calculate the complete noise conditional objective function. We use a uniform distribution on $\sigma \in [0,\sigma_{\mathrm{max}}]$ discretized at $H$ evenly spaced points starting at $0$ and terminating at $\frac{H-1}{H}\sigma_{\mathrm{max}}$. The value of $\sigma_{\mathrm{max}}$ is a parameter to be set. Without the prior,
\begin{equation}
 \mathcal{L}{} ~=~ \mathbb{E}_{p(\sigma)}[\mathcal{L}_{\sigma}] ~=~ \tfrac{1}{H}\sum\nolimits_{h=1}^{H} \mathcal{L}_{{\frac{h-1}{H}\sigma_\mathrm{max}}}.
\end{equation}
The gradient of $\mathcal{L}_\sigma$ for each value of $\sigma$ is
\begin{eqnarray}
    &&\hspace{-20pt}\nabla_{\theta}\mathcal{L}_{\sigma} = - \tfrac{1}{\gamma_\sigma^2}\|Z\|^2 \odot\sum\nolimits_{i=1}^N\mathbb{E}[\phi_\sigma(x_i+\xi_i)]\\
    &&\quad~\,\,\,  -  \tfrac{1}{\gamma_\sigma}\sum\nolimits_{i=1}^N\mathbb{E}[\phi_\sigma'(x_i+\xi_i)]\odot Z\Sigma_\sigma^{-1}(x_i-\mu)\nonumber\\
    && \quad~\,\,\, - \tfrac{1}{\gamma_\sigma}\sum\nolimits_{i=1}^N \mathbb{E}\left[( Z\Sigma_\sigma^{-1}\xi_i)\odot\phi_\sigma'(x_i+\xi_i)\right]\nonumber\\
        &&~~+ \Big( \tfrac{1}{\gamma_\sigma^2} ZZ^\top \odot\, \sum\nolimits_{i=1}^N\mathbb{E}[\phi_\sigma'(x_i+\xi_i)\phi_\sigma'(x_i+\xi_i)^\top]\Big)\theta .\nonumber
\end{eqnarray}
These terms parallel Equation \ref{eq.nablabasic} for the noise-free case. Using the expectations in Table \ref{tab.equations} and including the Gaussian prior regularization of $\theta$ with parameter $\lambda$, the closed form solution for $\theta$ that minimizes $\mathcal{L} + \frac{\lambda}{2}\theta^\top\theta$ is given in Algorithm \ref{alg:inference2}. We observe that Algorithm \ref{alg:inference} is the special case when $\sigma \in \{0\}$.

\subsection{Fisher Variational Predictive Distribution}\label{sec.vi}
Noise conditional score models overcome uncertainty in low data regions by adding noise to the data. As an alternative, we propose a Bayesian formulation that models uncertainty in $\theta$ and constructs the predictive distribution $p(x|x_1,\dots,x_N)$, with the same goal of eliminating high density artifacts in low information regions. Since the log-linear form of the TGP score uniquely allows for a predictive distribution to be approximated in closed form, it provides an appealing alternative here, though noise conditional models remain much easier to work with for deeper score functions.

First, we observe that the fact that Fisher divergence is quadratic in $\theta$, combined with the Gaussian prior on $\theta$, suggests approximating the posterior of $\theta$ with a Gaussian. Since the Fisher divergence does not correspond to a log likelihood, where $\ln q_\theta(x)$ would be expected instead, using it for variational inference stretches the Bayesian framework. Nevertheless, we note a tractable ELBO-type function suggests itself for optimization,
\begin{equation}\label{eq.elbo}
    \mathcal{L}(q(\theta)) = \mathbb{E}_{q(\theta)}\hspace{-3pt}\left[-\frac{1}{2\eta}\mathbb{E}_{p(x)}\Big\|\nabla_x \ln \frac{p(x)}{q_{\theta}(x)}\Big\|^2 + \ln \frac{p(\theta)}{q(\theta)}\right].\nonumber
\end{equation}
We see the log likelihood of the typical ELBO is replaced with the negative Fisher divergence. We also include an additional variational tempering (VT) parameter $\eta$, originally proposed by Mandt et al. \cite{mandt2016variational} to anneal the ELBO for better local optimal learning of $q(\theta)$. In VT, $\eta$ decreases to $1$ as a function of iteration to ensure optimization of the desired ELBO; since we will be able to find the optimal $q(\theta)$ in closed form, we instead use $\eta$ to perform a necessary regularization to balance the Fisher and KL divergence terms in $\mathcal{L}$.

Following the VI theory \cite{bishop2006pattern}, the $q(\theta)$ that maximizes $\mathcal{L}$ is 
\begin{equation}
q(\theta) \propto \exp \left[-\frac{1}{2\eta}\mathbb{E}_{p(x)}\Big\|\nabla_x \ln \frac{p(x)}{q_{\theta}(x)}\Big\|^2\right]p(\theta).
\end{equation}
Because the Fisher divergence consists of terms linear and quadratic in $\theta$, as shown in Section \ref{sec.basic}, the solution to $q(\theta)$ is the following multivariate Gaussian distribution,
\begin{align}\label{eq.posterior}
\qquad q(\theta) &= \mathcal{N}(\theta\,|\,\widehat{\mu},\,\widehat{\Sigma})\\
    \widehat{\Sigma} &=  \big(\lambda I + \tfrac{1}{\gamma^2\eta}ZZ^\top \odot \Phi'\big)^{-1}\nonumber\\
    \widehat{\mu} &=  \left(\lambda\gamma^2\eta I + ZZ^\top \odot \Phi'\right)^{-1}\left(\gamma\psi' + \|Z\|^2\odot\psi\right).\nonumber
\end{align}
The definitions of $\Phi'$, $\psi$ and $\psi'$ are taken from Algorithm \ref{alg:inference}. We observe that the equivalence between $\widehat{\mu}$ above and the Fisher divergence solution given in Algorithm \ref{alg:inference} parallels that between ridge regression and Bayesian linear regression. In effect, the variational Fisher distribution $q(\theta)$ adds Gaussian uncertainty to the $\theta_{\mathrm{FD}}$ point estimate.

Using the posterior approximation $q(\theta)$, we can approximate the predictive distribution. Let data $X = \{x_1,\dots,x_N\}$. The variational predictive distribution is
$$ q(x|X) = \int q_\theta(x)q(\theta)\,d\theta$$
and we recall that $q_{\theta}(x) \propto \exp\lbrace{\theta^\top \phi(x)\rbrace} \mathcal{N}(x|\mu,\Sigma)$. The integral is intractable because $\theta$ appears in the also-intractable denominator of $q_\theta(x)$. We introduce the following MGF approximation of the normalizing constant of $q_\theta(x)$,
$$\mathbb{E}_{x\sim\mathcal{N}(x|\mu,\Sigma)}\big[\exp\{\theta^\top \phi(x)\}\big] \approx \exp\{\theta^\top\mu_\phi + \tfrac{1}{2}\theta^\top \Sigma_\phi \theta\},$$
\begin{equation}\label{eqn.mgfapprox}
    \mu_\phi = \mathbb{E}[\phi(x)],\quad \Sigma_\phi = \mathbb{E}[\phi(x)\phi(x)^\top] - \mu_\phi\mu_\phi^\top.
\end{equation}
Using a similar derivation as in Section \ref{sec.noise} for the noise conditional model, we can show that
\begin{align}
\mu_\phi(s) =&~ \exp\{ - \tfrac{1}{2\gamma^2}z_s^\top\Sigma z_s\}\phi_s(\mu)\label{eq.muphi}\\
\Sigma_\phi =&~ \tfrac{1}{2}(\Delta^-+\Delta^+)\odot\phi(\mu)\phi(\mu)^\top\label{eq.sigphi}\\
+&~ \tfrac{1}{2}(\Delta^--\Delta^+)\odot\phi'(\mu)\phi'(\mu)^\top - \mu_\phi\mu_\phi^\top\nonumber    \\
\Delta_{ss'}^+ :=&~ \exp \{-\tfrac{1}{2\gamma^2}(z_s+z_{s'})^\top\Sigma(z_s+z_{s'})\}\nonumber\\
\Delta_{ss'}^- :=&~ \exp\{ - \tfrac{1}{2\gamma^2}(z_s-z_{s'})^\top\Sigma(z_s-z_{s'})\}.\nonumber
\end{align}
With this approximation, the predictive distribution becomes
$$ q(x|X) \approx \mathcal{N}(x|\mu,\Sigma)\int\mathrm{e}^{\theta^\top(\phi(x)-\mu_\phi)- \frac{1}{2}\theta^\top\Sigma_\phi\theta}\mathcal{N}(\theta|\widehat{\mu},\widehat{\Sigma})\,d\theta.$$
The integral can be solved and the approximate variational predictive distribution is
$$ q(x|X) \approx \mathcal{N}(x|\mu,\Sigma)\mathrm{e}^{\frac{1}{2}\phi(x)^\top\mathbf{M}^{-1}\phi(x) - \phi(x)^\top \mathbf{M}^{-1}\mathbf{m}},$$
\begin{equation}\label{eq.predictive}
\mathbf{m} = \mu_\phi - \widehat{\Sigma}^{-1}\widehat{\mu},\quad    \mathbf{M} = \Sigma_\phi + \widehat{\Sigma}^{-1}.
\end{equation}
We summarize this \textit{Fisher variational predictive distribution} (FVPD) in Algorithm \ref{alg:inference3}.

\begin{algorithm}[t]
\caption{Fisher variational predictive distribution for TGP}\label{alg:inference3}
\begin{algorithmic}[1]
\Require Data $x \in\mathbb{R}^d$, kernel width $\gamma > 0$, param $\lambda >0$, $S\in\mathbb{Z}_+$, base params $\mu\in\mathbb{R}^d$, $\Sigma \in \mathbb{S}_{++}^{d}$, variational tempering param $\eta > 0$ (we set $\eta = 1/\gamma^2$)
\State Run  Algorithm \ref{alg:inference} steps 1 through 4. 
\State Calculate $\widehat{\mu}$, $\widehat{\Sigma}$, $\mu_\phi$, $\Sigma_\phi$ (Eqs.\ \ref{eq.posterior}, \ref{eq.muphi} \& \ref{eq.sigphi})
\State Construct the approximate predictive distribution
$$  p(x|X) \propto \mathcal{N}(x|\mu,\Sigma)\mathrm{e}^{\frac{1}{2}\phi(x)^\top\mathbf{M}^{-1}\phi(x) - \phi(x)^\top \mathbf{M}^{-1}\mathbf{m}}$$
$$\mathbf{m} = \mu_\phi - \widehat{\Sigma}^{-1}\widehat{\mu},\quad    \mathbf{M} = \Sigma_\phi + \widehat{\Sigma}^{-1}$$
\Return unscaled $p(x|X)$ with params $\mu$, $\Sigma$, $\textbf{m}$, $\textbf{M}$, $Z$, $c$
\end{algorithmic}
\end{algorithm}

\begin{table*}[htp!]
\centering
\resizebox{1\textwidth}{!}{
\begin{tabular}{c|ccccccccccc} 
\hline\hline
\textbf{Algorithm} &\textbf{Bikeshare} & \textbf{MIMIC2} & \textbf{MAGIC} & \textbf{HIGGS}  & \textbf{Credit} & \textbf{Power}& \textbf{GasMod}&\textbf{GasMix}&\textbf{HARTH}& \textbf{Twitter}\\
\hline
MAP & \textbf{2.25}$\,|\,$\textbf{3.60}& 1.78$\,|\,$2.78&7.05$\,|\,$14.1& 0.67$\,|\,$1.21&2.12$\,|\,$4.62 & 11.7$\,|\,$23.1 &7.17$\,|\,$11.3&\textbf{7.80}$\,|\,$13.7&10.1$\,|\,$21.4& 6.00$\,|\,$\textbf{12.0}\\
FD & \textbf{2.25}$\,|\,$3.82&1.78$\,|\,$2.84& 5.45$\,|\,$12.1 &\textbf{0.60}$\,|\,$\textbf{1.03}& 2.10$\,|\,$4.56 & 11.3$\,|\,$23.0&\textbf{6.79}$\,|\,$11.1&8.23$\,|\,$14.0&10.1$\,|\,$21.2& 6.15$\,|\,$12.4 \\ 
NCFD & 2.29$\,|\,$3.81&1.79$\,|\,$2.88& 5.44$\,|\,$\textbf{11.9} &0.66$\,|\,$1.21& 2.05$\,|\,$\textbf{4.42} & \textbf{11.0}$\,|\,$\textbf{21.4}&7.05$\,|\,$11.9 &8.49$\,|\,$13.9&10.1$\,|\,$21.9& \textbf{5.72}$\,|\,$13.2 \\
FVPD & 2.28$\,|\,$3.70&\textbf{1.70}$\,|\,$\textbf{2.76}& \textbf{5.34}$\,|\,$\textbf{11.9} & 0.65$\,|\,$1.19&\textbf{2.01}$\,|\,$4.46 & 11.4$\,|\,$23.0&6.91$\,|\,$\textbf{10.8}&8.43$\,|\,$\textbf{13.5}&10.1$\,|\,$\textbf{21.1}& 5.87$\,|\,$12.3 \\
\hline\hline
MAP & 2.78$\,|\,$4.74& 2.17$\,|\,$3.49& 7.44$\,|\,$16.1& 0.73$\,|\,$1.38& 2.32$\,|\,$5.06& 12.7$\,|\,$26.0&7.71$\,|\,$12.9& \textbf{8.39}$\,|\,$\textbf{15.2}& 10.3$\,|\,$21.8& \textbf{6.38}$\,|\,$\textbf{12.6}\\
FD & 2.73$\,|\,$4.67& 1.98$\,|\,$3.13& 6.04$\,|\,$13.8& \textbf{0.67}$\,|\,$\textbf{1.24}& 2.29$\,|\,$4.93& 12.2$\,|\,$25.4&7.33$\,|\,$12.8& 8.70$\,|\,$16.1& \textbf{10.2}$\,|\,$21.7& 6.77$\,|\,$13.3\\
NCFD & 2.70$\,|\,$4.57& 2.01$\,|\,$3.23 & 5.94$\,|\,$\textbf{13.3} & 0.74$\,|\,$1.43 & 2.24$\,|\,$\textbf{4.83}& \textbf{11.9}$\,|\,$\textbf{24.6}& 7.43$\,|\,$13.0& 8.85$\,|\,$16.1& 10.3$\,|\,$21.7& 6.44$\,|\,$13.9\\
FVPD & \textbf{2.62}$\,|\,$\textbf{4.47}& \textbf{1.96}$\,|\,$\textbf{3.12}& \textbf{5.88}$\,|\,$13.5& 0.72$\,|\,$1.37& \textbf{2.23}$\,|\,$4.85& 12.0$\,|\,$25.0&\textbf{7.32}$\,|\,$\textbf{12.2}&  8.64$\,|\,$\textbf{15.2}& \textbf{10.2}$\,|\,$\textbf{21.4}& 6.40$\,|\,$12.8\\
\hline
\hline
\end{tabular}}
\caption{Kolmogorov-Smirnov test scores and Wasserstein distances (KS$\,|\,$WD) between the empirical distributions and the marginal along random one dimensional subspaces. All values are shown on scale $\times 10^{-2}$. The top shows the median and the bottom the mean of the KS/WD values over uniform projection distribution using a 500-sample Monte Carlo approximation. } \label{tab.quantitative}
\end{table*}

We can observe some properties of (\ref{eq.predictive}) to connect it with the base FD Algorithm \ref{alg:inference} and help guide setting the parameter $\eta$ that appears in $\widehat{\mu}$ and $\widehat{\Sigma}$. First, since the squared kernel width $\gamma^2$ may be very small, $\mathbf{M}^{-1}\mathbf{m} \approx -\widehat{\mu}$ and $\phi(x)^\top \mathbf{M}^{-1}\phi(x) \approx 0$ when $\eta=1$. The predictions of Algorithm \ref{alg:inference} and Algorithm \ref{alg:inference3} become nearly the same since $\widehat{\mu} = \theta_{\mathrm{FD}}$ and the covariance $\widehat{\Sigma}$ is small. In this scenario, we empirically observed almost identical results as Algorithm \ref{alg:inference} since a huge $1/\gamma^2$ scaling removes uncertainty in $\theta$. To eliminate this effect, the tempering parameter $\eta$ can act as a regularizer to balance the Fisher and KL divergence terms of the ELBO. We found that setting $\eta \propto 1/\gamma^2$ can remove the undesirable impact of $\gamma$ in (\ref{eq.posterior}).

\section{Experiments}\label{sec.experiments}
We conduct an empirical evaluation to demonstrate our proposed methods on lower dimensional density estimation problems. We test the our four inference algorithms for the GP-tiled density: MAP, Fisher Divergence (FD), Noise Conditional Fisher Divergence (NCFD), and Fisher Variational Predictive Distribution (FVPD).

\paragraph{Datasets} We evaluate the four algorithms in this paper on ten data sets from the UCI repository and four other illustrative data sets. These include Bikeshare, MIMIC2, MAGIC, HIGGS, Credit, Power, GasMod, GasMix, HARTH and Twitter.  Statistics for these data sets are given in Table \ref{table:datasets}. We also illustrate using 2D-projected MNIST, two dimensions of CA Housing (from UCI), and two simulated 2D data sets.

\paragraph{Parameter settings} We set $S=1000$ and $\lambda = 0.1$. We also follow ``Scott's Rule'' to set $\gamma = N^{-1/(d+4)}\sqrt{\mathrm{tr}(\mathbb{V}(x))/d}$ \cite{scott2015multivariate}. We set $\mu$ and $\Sigma$ in the base normal distribution to be their empirical values from the data. For the noise conditional model we set $\sigma_{\mathrm{max}} = \frac{1}{d}\sqrt{\mathrm{tr}(\mathbb{V}(x))}$ and $H=10$. For FVPD, we set $\eta = \frac{1}{\gamma^2}$. All values were obtained through rough parameter selection, but could be further optimized. While we discuss TGP in the context of a spherical kernel, we sample $z\sim\mathcal{N}(0,\Sigma_z)$, where $\Sigma_z = d\cdot\Sigma/\mathrm{tr}(\Sigma)$ to incorporate data correlation in the kernel function, with slight performance improvement.

\paragraph{Quantitative Result} We show a quantitative evaluation on 10 public benchmark data sets in Table \ref{tab.quantitative}. As a performance measure, we use the Kolmogorov-Smirnov (KS) test and the Wasserstein distance (WD) to quantify similarity between the learned and empirical distributions. Let $\mathcal{P}_N$ be the empirical CDF of data and $Q$ be the CDF of a learned distribution on the same data, both in $\mathbb{R}$. KS and WD measure discrepancy between these CDFs in one dimension as
\begin{eqnarray}
    \mathrm{KS}(\mathcal{P}_N,Q) &=& \textstyle\sup_x |\mathcal{P}_N(x) - Q(x)|,\nonumber\\  
    \mathrm{WD}(\mathcal{P}_N,Q) &=&\textstyle \int_{\mathbb{R}} |\mathcal{P}_N(x) - Q(x)|\,dx.\nonumber
\end{eqnarray}

\renewcommand{\arraystretch}{1.2}{
\begin{table}[bhp!]
\resizebox{1\columnwidth}{!}{
\begin{tabular}{lrr|cccc}
 \hline\hline
 \textbf{Dataset} & \textbf{N} & \textbf{D} & \textbf{MAP}  & \textbf{FD} & \textbf{NCFD} & \textbf{FVPD}  \\ 
 \hline
  Bikeshare    &11,122&12& 5.7m & 0.18s  & 3.6s & 0.18s  \\
 MIMIC2     &15,684&17& 5.7m  & 0.31s & 9.2s & 0.31s  \\
  MAGIC        &19,020&10& 5.7m  & 0.28s  & 5.1s & 0.28s  \\
 HIGGS        &50,000&28& 5.7m & 0.56s & 7.3s & 0.56s  \\
 Credit       &182,276 & 30 & 5.7m & 2.3s  & 30s & 2.3s    \\
 Power        & 2,049,280&7& 6m &22s&4.5m&22s \\
  GasMod       & 3,843,160& 18 & 6.4m & 43s & 9.8m& 43s \\
 GasMix       &4,208,261&16& 6.5m & 45s& 10m&45s\\
 HARTH        &6,461,328& 6 & 7m &68s& 13m &68s\\
 Twitter      &14,262,515 & 3 & 8m & 2.5m & 33m & 2.5m \\
 \hline\hline
\end{tabular}
}
\caption{Data statistics and total running times.}\label{table:datasets}
\end{table} 
}

\begin{figure*}[tp!]
    \centering
    \subfloat[MAP (Algorithm 1)]{\includegraphics[width=.25\textwidth]{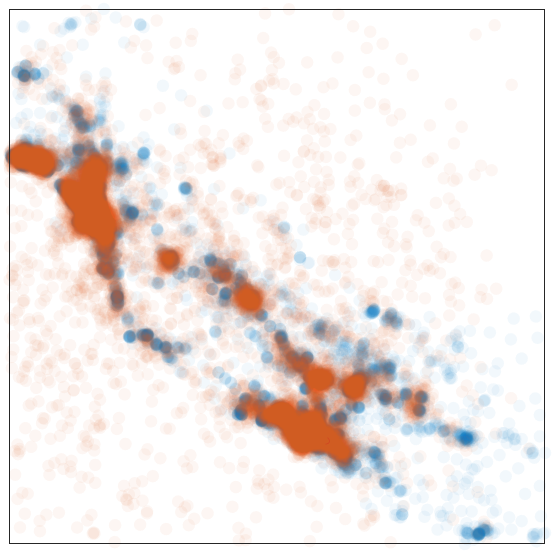}}   
    \subfloat[FD (Algorithm 2)]{\includegraphics[width=.25\textwidth]{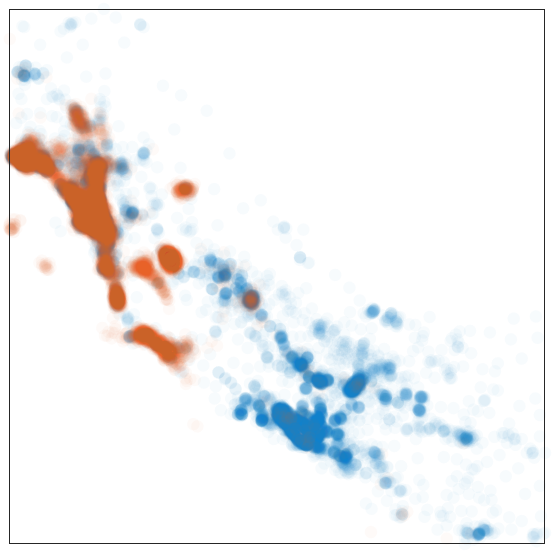}}
    \subfloat[NCFD (Algorithm 3)]{\includegraphics[width=.25\textwidth]{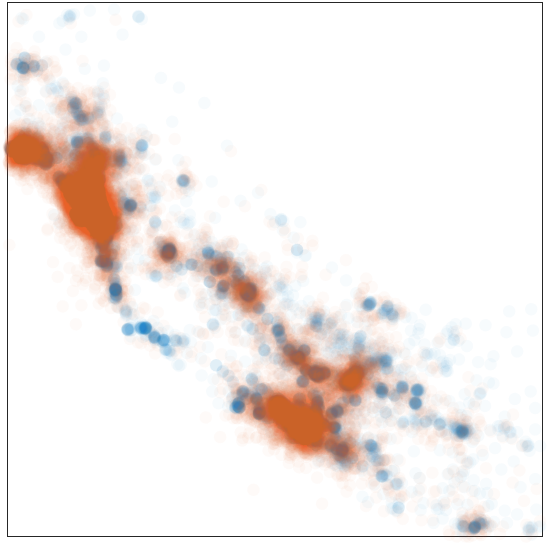}}
    \subfloat[FVPD (Algorithm 4)]{\includegraphics[width=.25\textwidth]{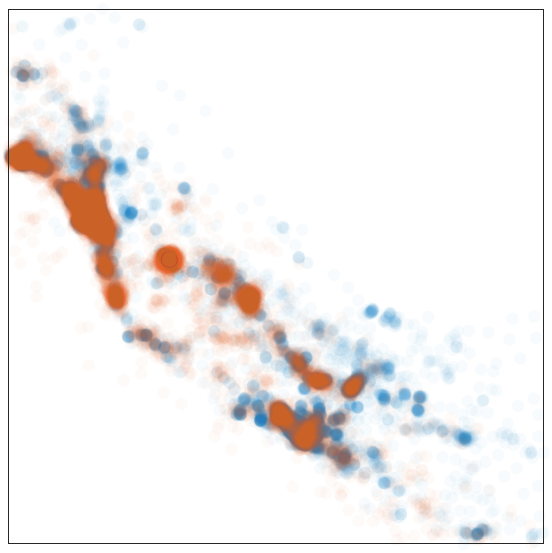}}\\
    \subfloat[KDE (RFF, $S=100K$)]{\includegraphics[width=.25\textwidth]{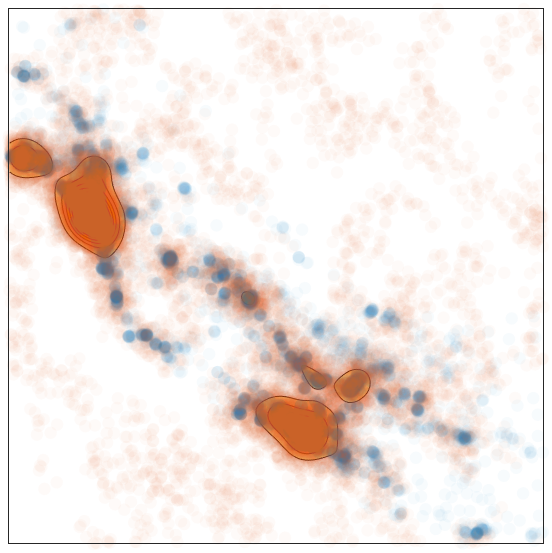}}
    \subfloat[KDE ($S=\infty$)]{\includegraphics[width=.25\textwidth]{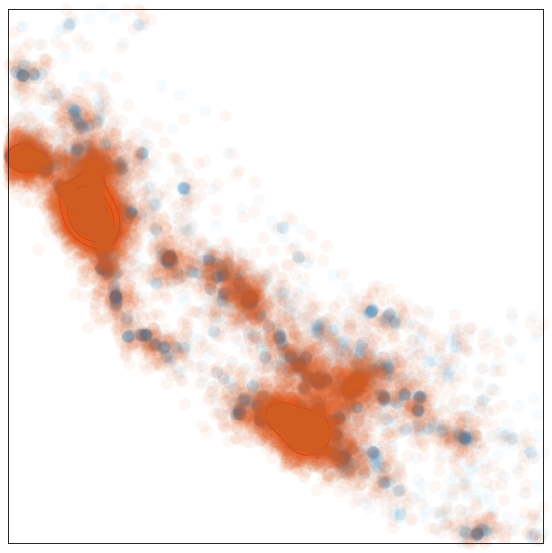}}\qquad
    \subfloat[Data]{\includegraphics[width=.25\textwidth]{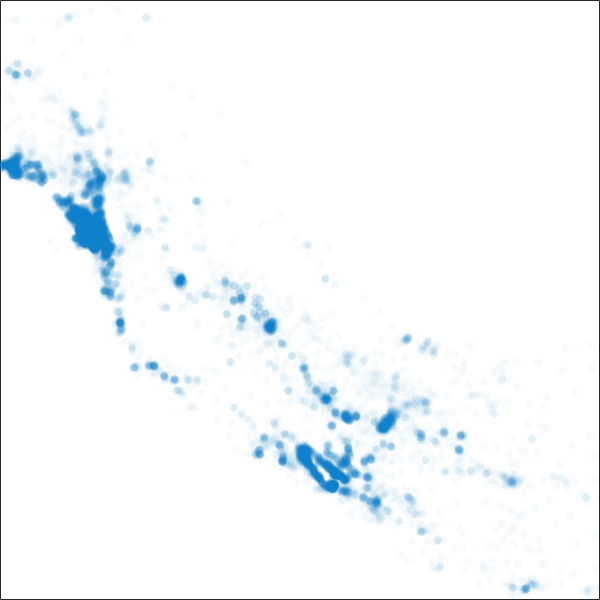}}
    \caption{Samples (orange) shown over data (blue) generated for (a-d) Algorithms 1-4, and (e,f) kernel density estimation (see text for discussion). We set $S=1000$ for (a)-(d) and $S=100K$ for (e), $\lambda = 10$ for (a-d). All methods use the same kernel width $\gamma$.}
    \label{fig:CA_samps}
\end{figure*}

\begin{figure*}[h!]
    \includegraphics[width=.2\textwidth]{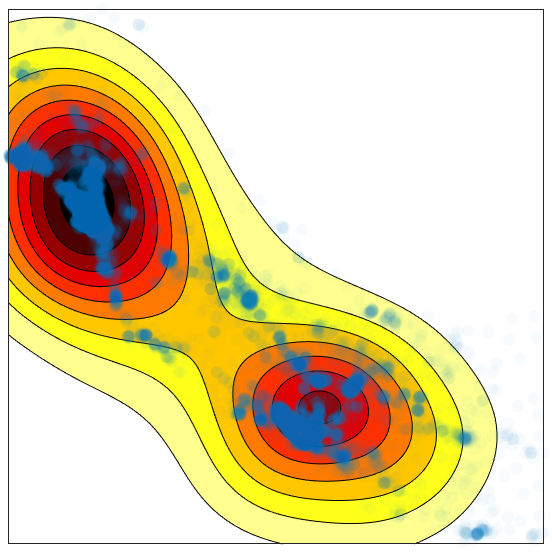}\hspace{-3pt}
    \includegraphics[width=.2\textwidth]{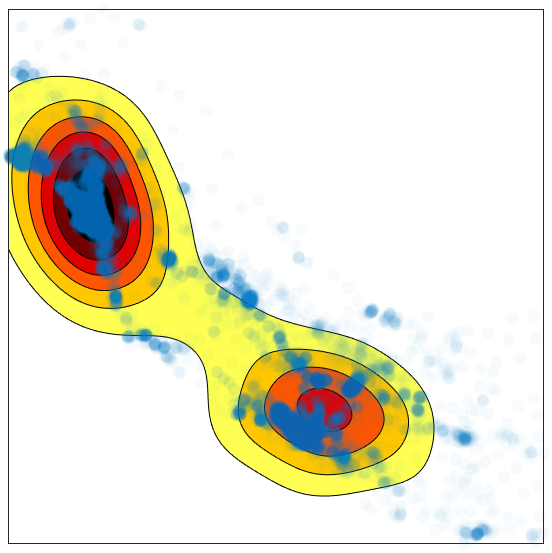}\hspace{-3pt}
    \includegraphics[width=.2\textwidth]{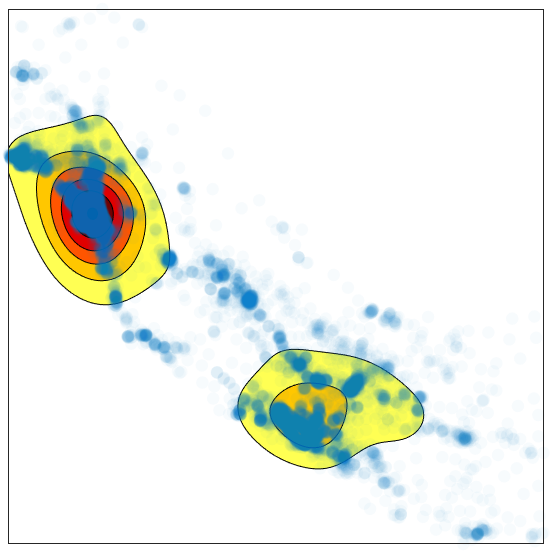}\hspace{-3pt}
    \includegraphics[width=.2\textwidth]{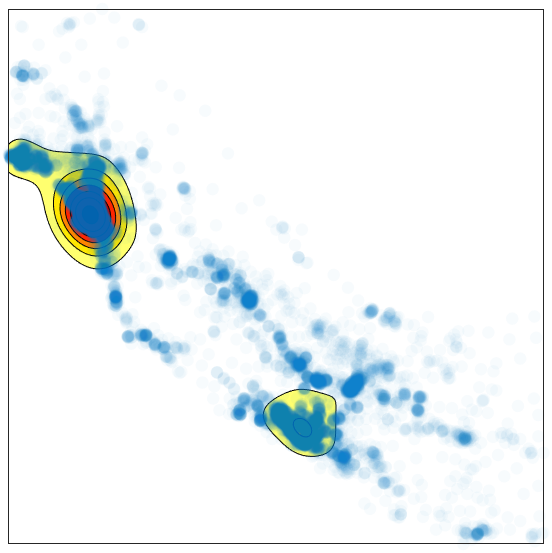}\hspace{-3pt}
    \includegraphics[width=.2\textwidth]{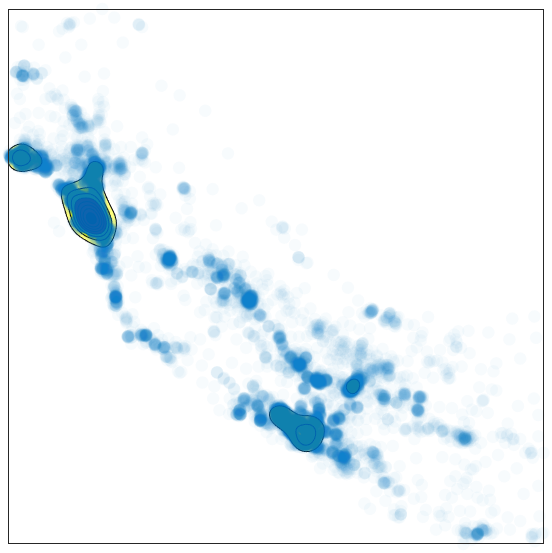}\hspace{-3pt}
    \caption{Algorithm 3 example. Noise conditional density contours on CA House data for decreasing $\sigma$. For better visualization, Figure \ref{fig:CA_samps} shows samples from the final RHS density contour (NCFD), along with samples from the other learned methods.}
    \label{fig:NCFD_CA}
\end{figure*}

\begin{figure}[bh!]
\vspace{-25pt}
    \subfloat{\includegraphics[height=.75in,width=0.2\columnwidth]{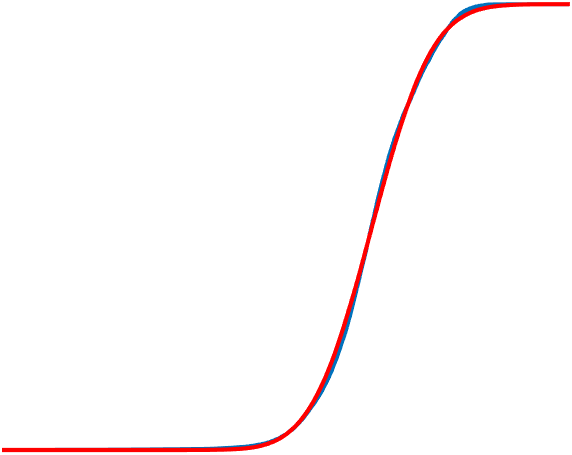}}
    \subfloat{\includegraphics[height=.75in,width=0.2\columnwidth]{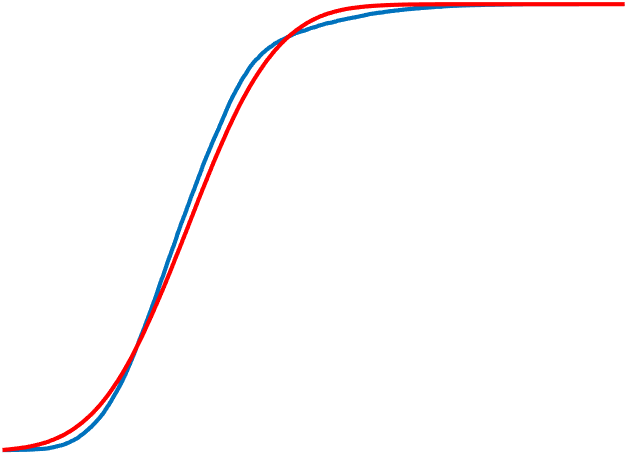}}
    \subfloat{\includegraphics[height=.75in,width=0.2\columnwidth]{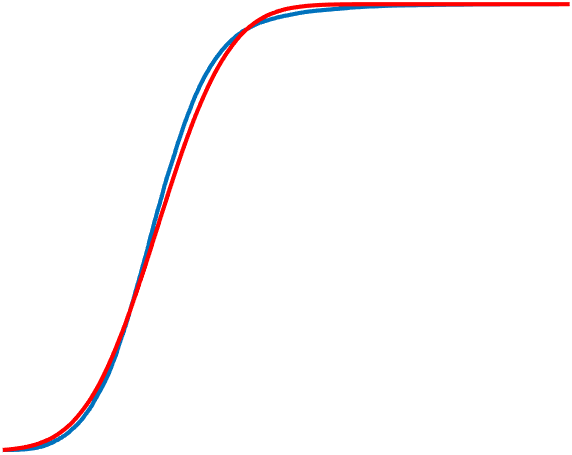}}
    \subfloat{\includegraphics[height=.75in,width=0.2\columnwidth]{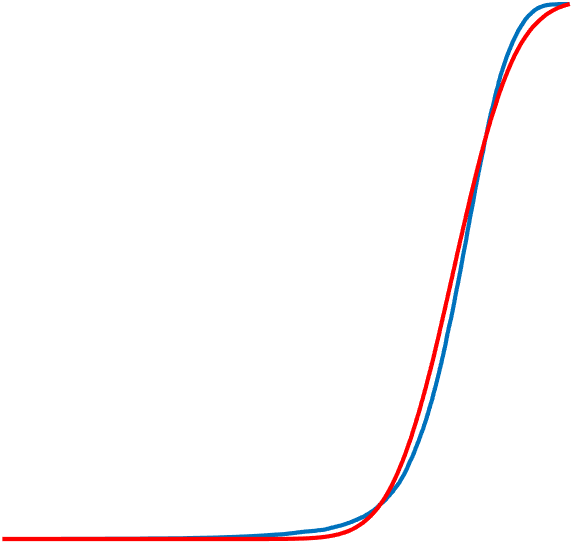}}
    \subfloat{\includegraphics[height=.75in,width=0.2\columnwidth]{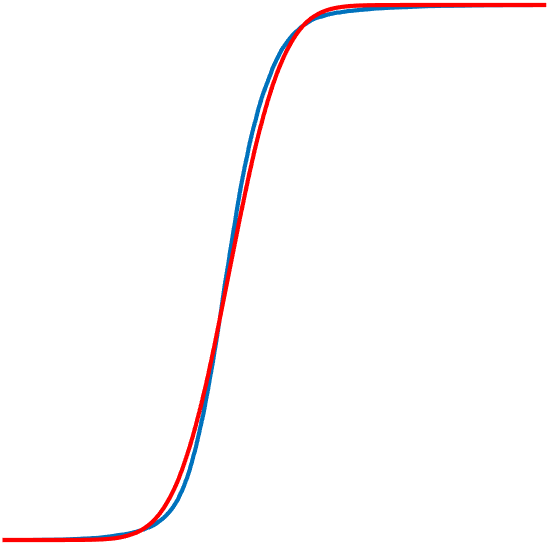}}\\
    \subfloat{\includegraphics[height=.75in,width=0.2\columnwidth]{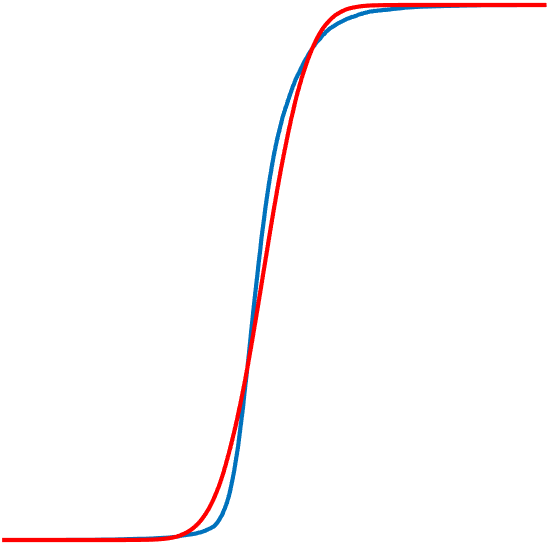}}
    \subfloat{\includegraphics[height=.75in,width=0.2\columnwidth]{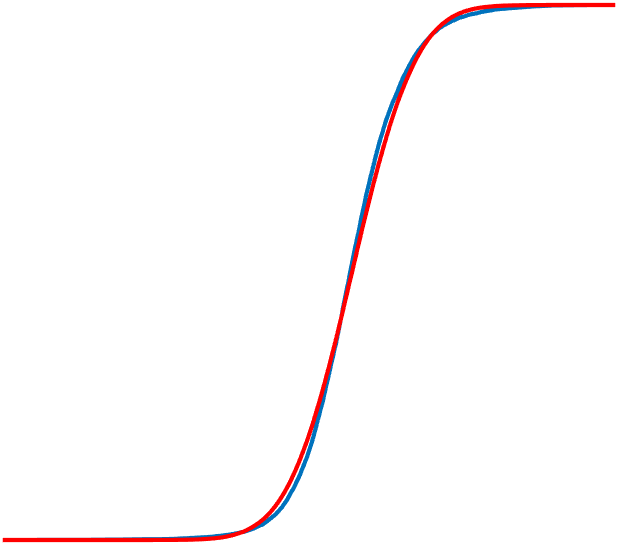}}
    \subfloat{\includegraphics[height=.75in,width=0.2\columnwidth]{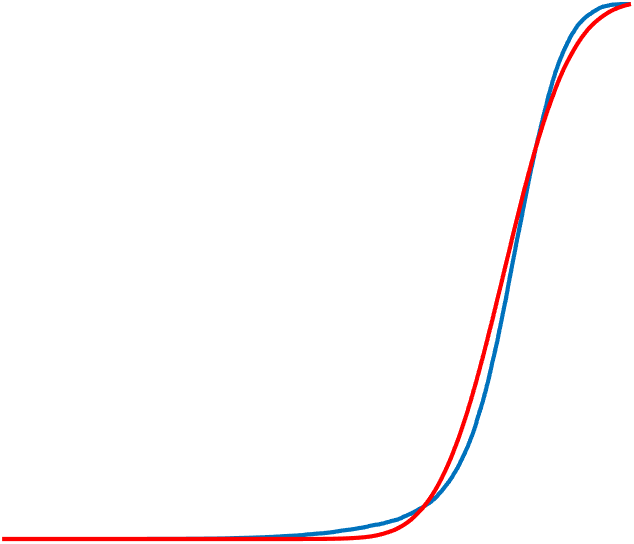}}
    \subfloat{\includegraphics[height=.75in,width=0.2\columnwidth]{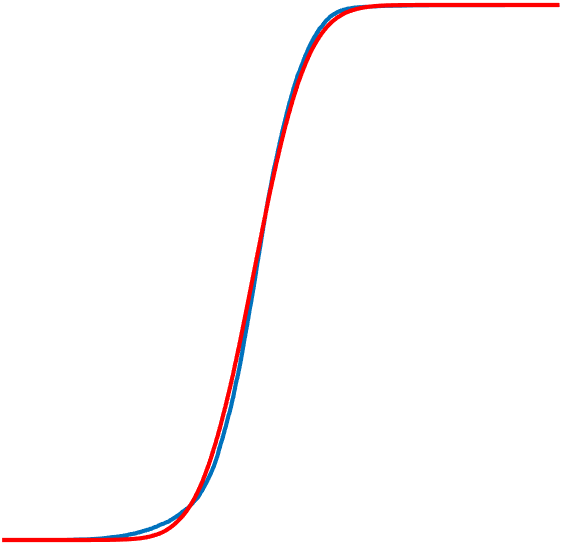}}
    \subfloat{\includegraphics[height=.75in,width=0.2\columnwidth]{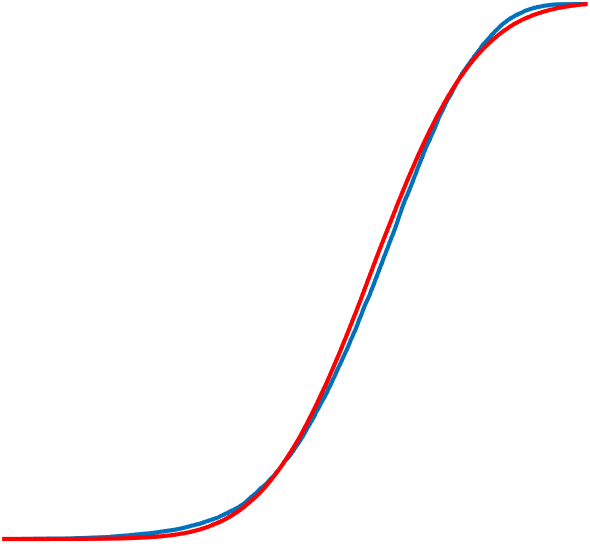}}
    \caption{Example CDFs of projections of MAGIC data used to calculate KS and WD. Data (blue) vs FVPD (red). Each plot corresponds to a different random $\boldsymbol{\mathrm{v}}$.}\label{fig.KSexamples}
\end{figure}

\begin{figure*}[t!]
\captionsetup[subfigure]{labelformat=empty}
\centering
   \subfloat{\includegraphics[width=0.195\textwidth]{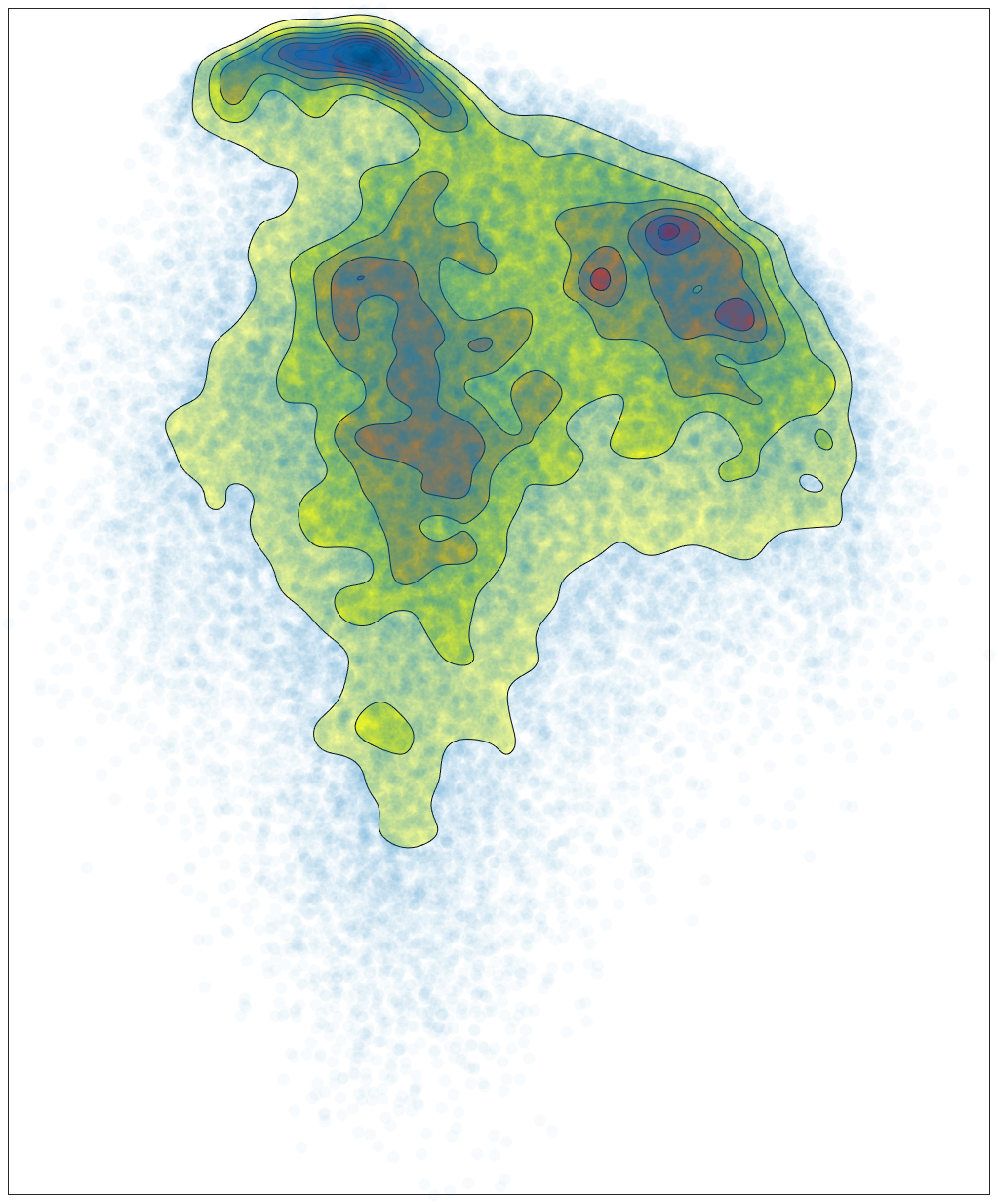}}
    \subfloat{\includegraphics[width=0.195\textwidth]{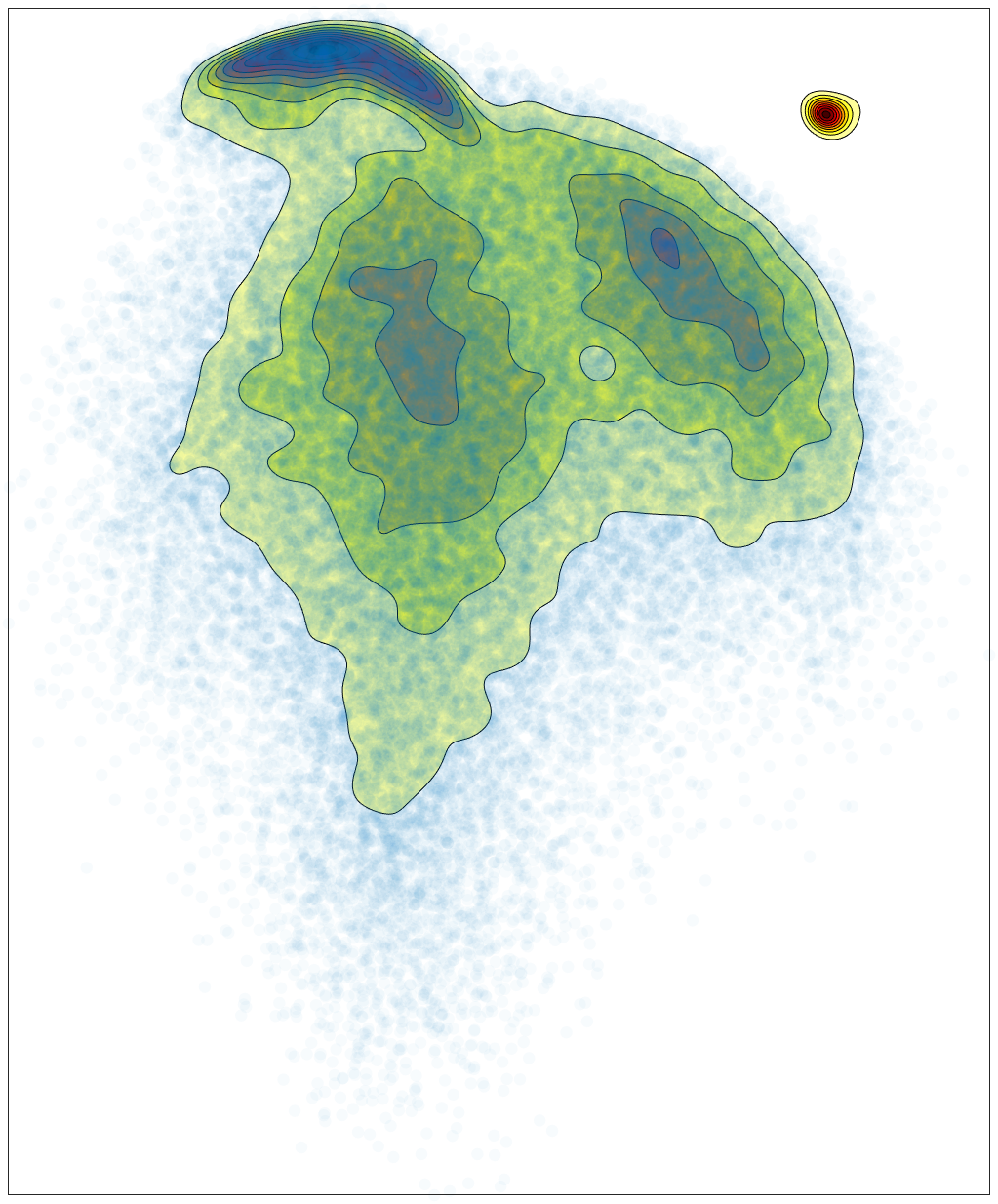}}
    \subfloat{\includegraphics[width=0.195\textwidth]{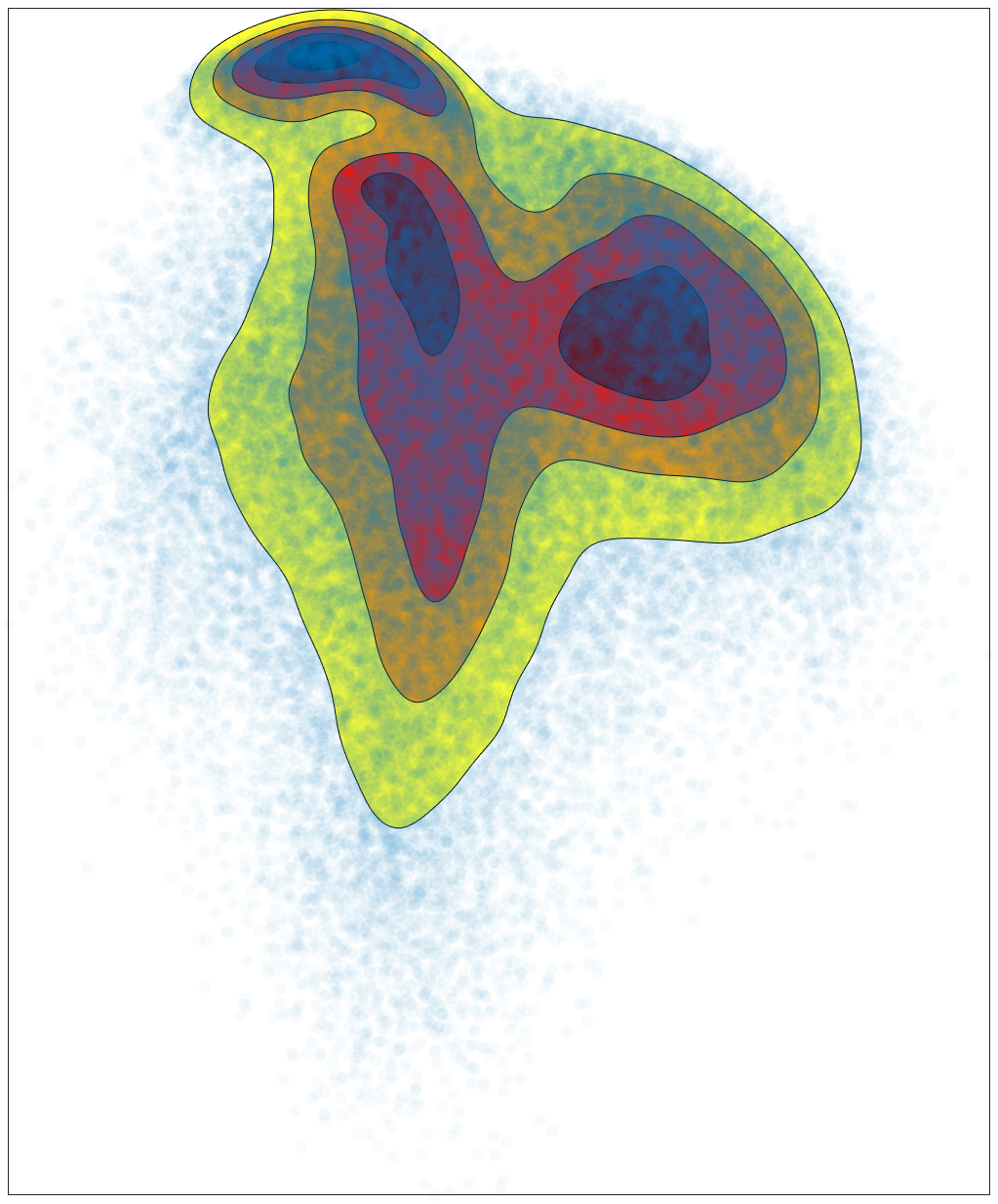}}
    \subfloat{\includegraphics[width=0.195\textwidth]{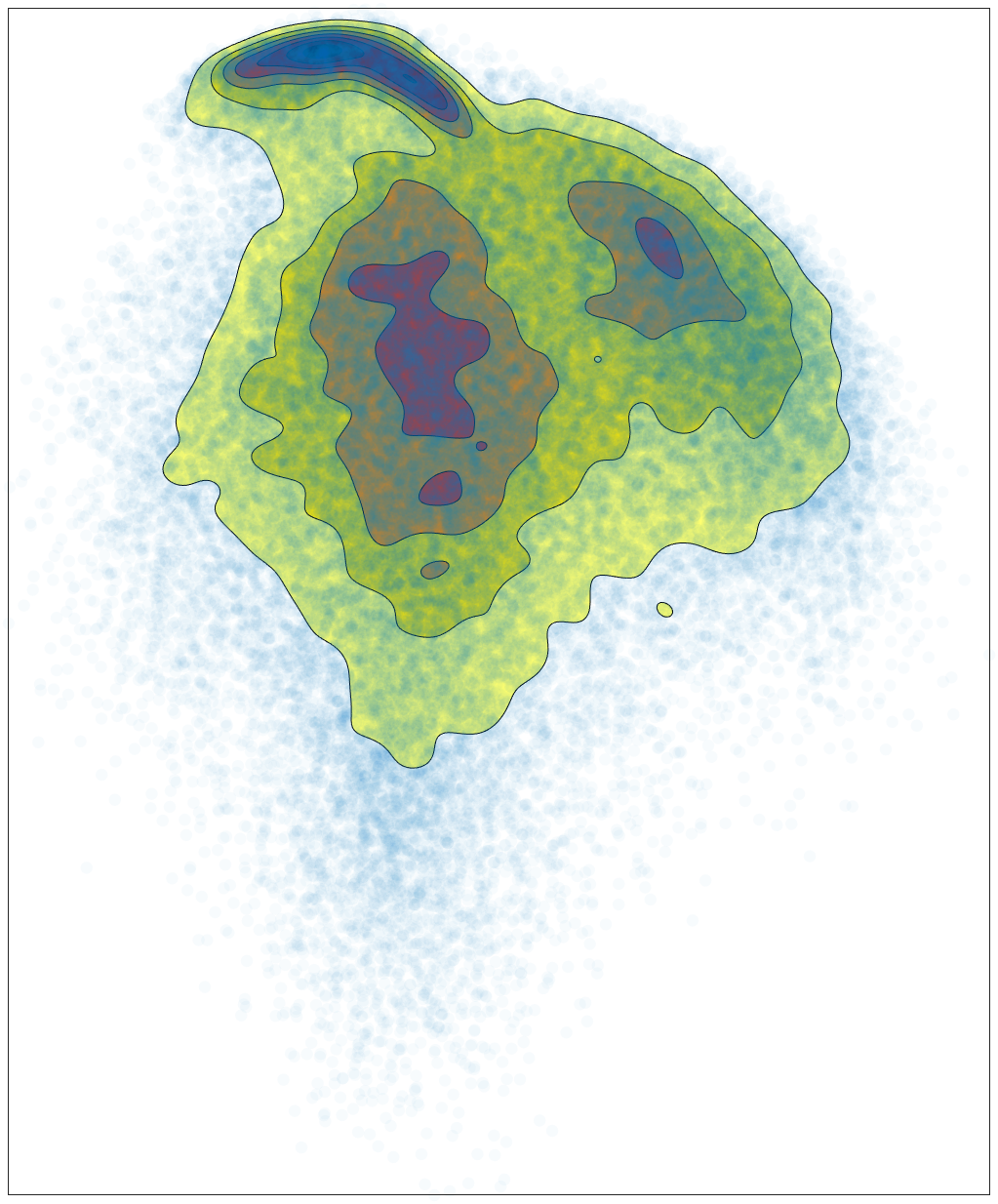}}\quad
   \subfloat{\includegraphics[width=0.195\textwidth]{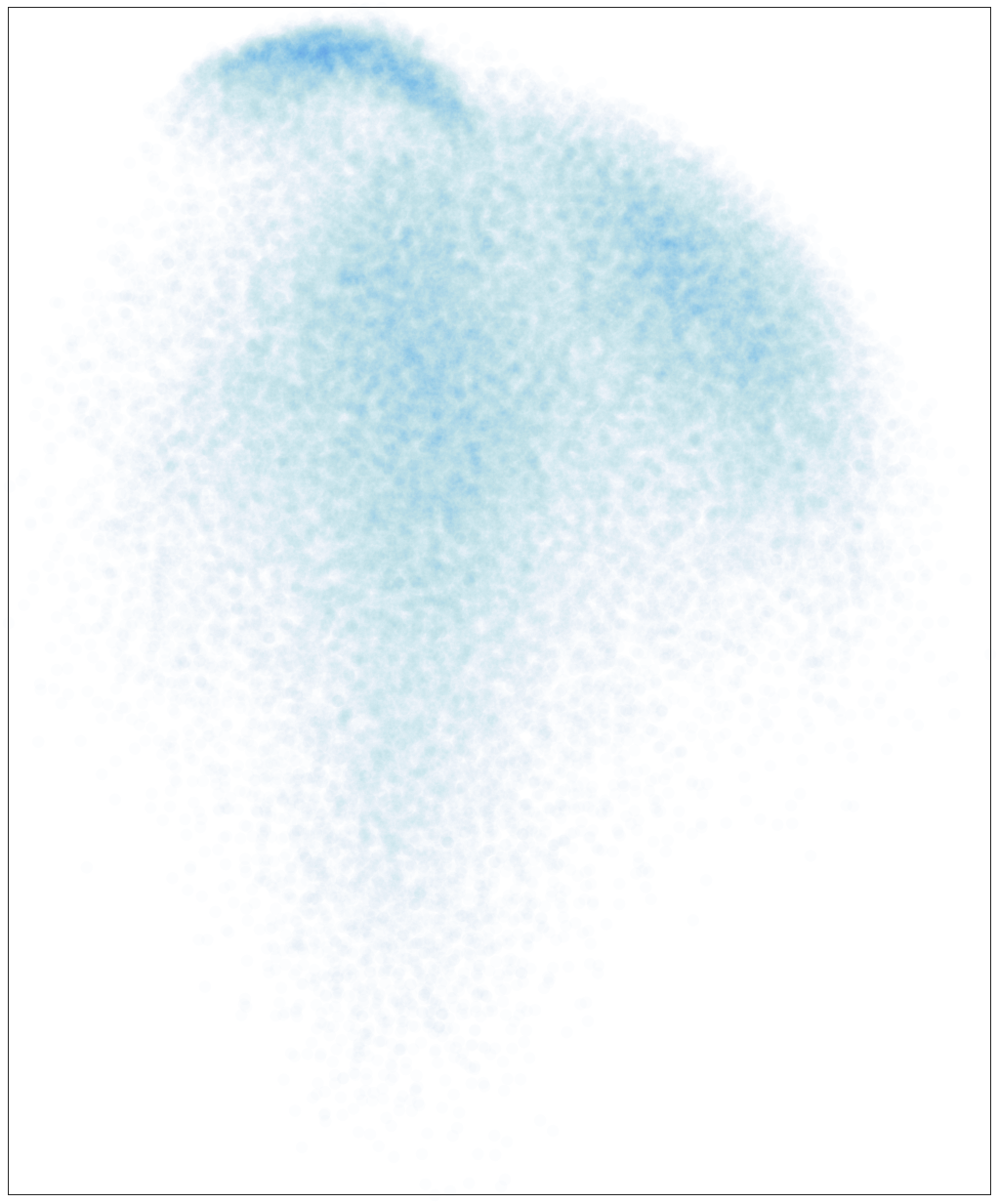}}\\\vspace{-10pt}
    \subfloat{\includegraphics[width=0.195\textwidth]{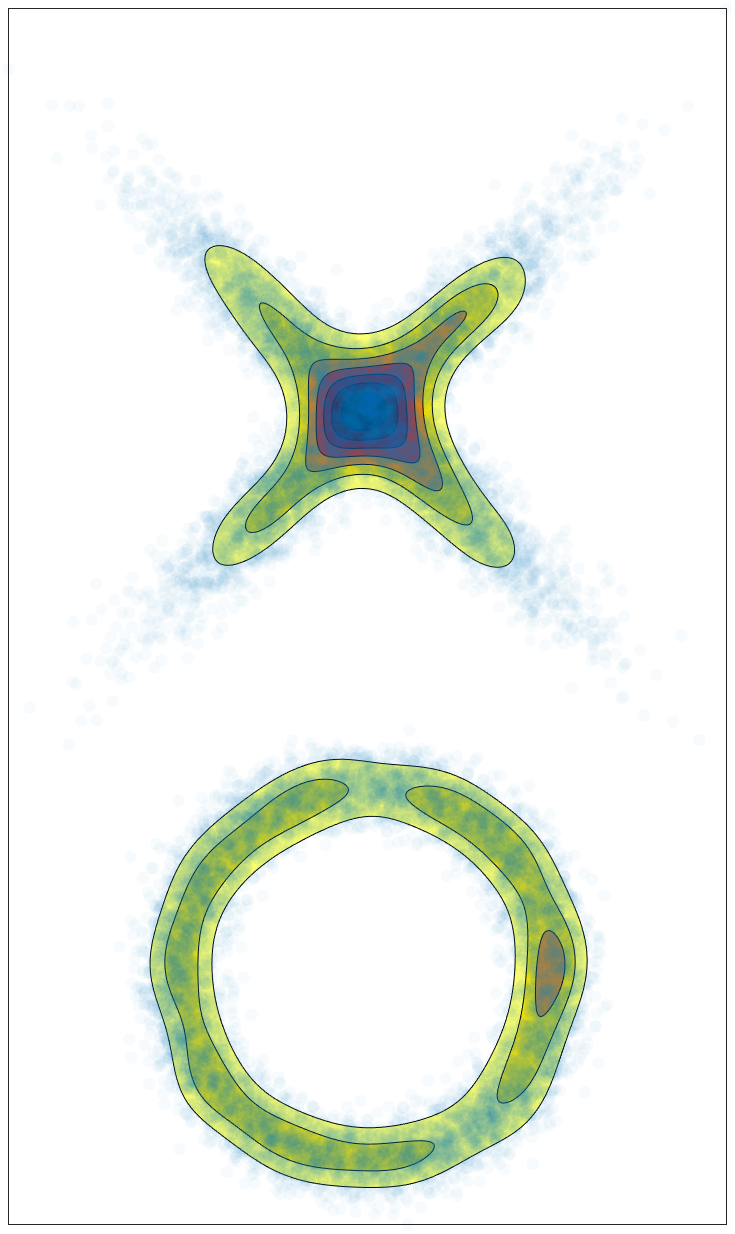}}
    \subfloat{\includegraphics[width=0.195\textwidth]{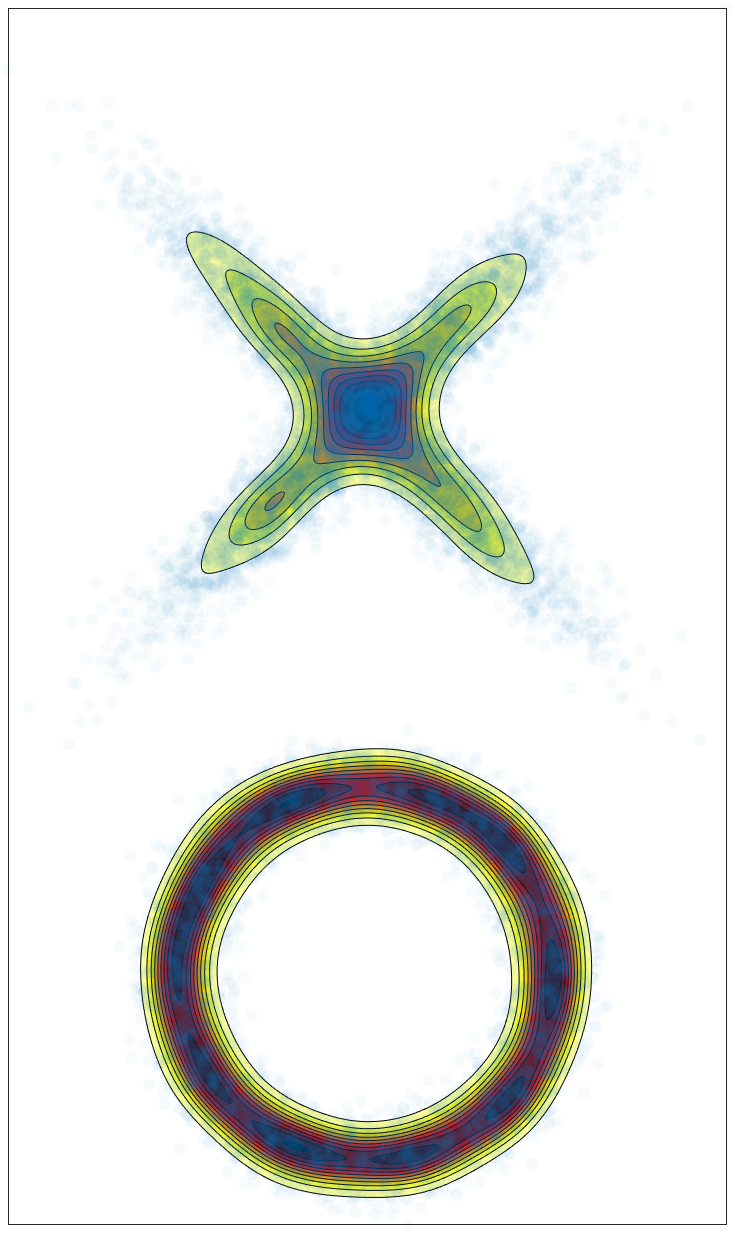}}
        \subfloat{\includegraphics[width=0.195\textwidth]{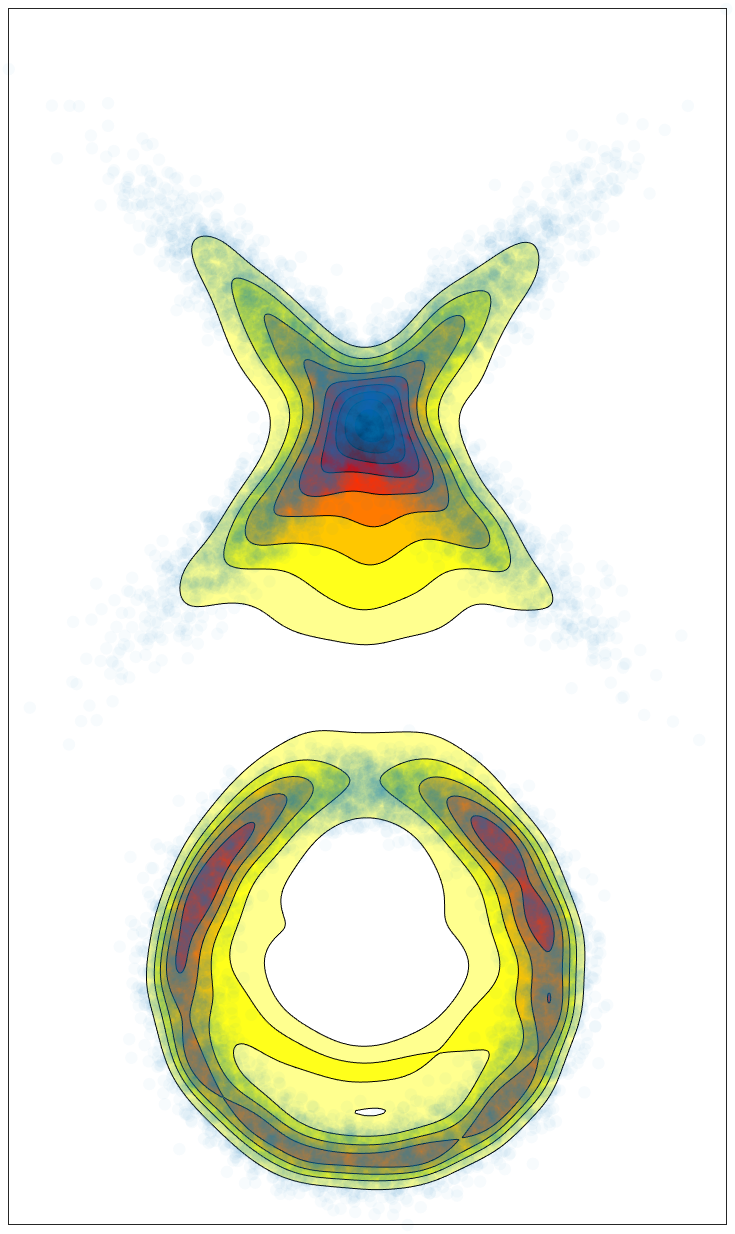}}
    \subfloat{\includegraphics[width=0.195\textwidth]{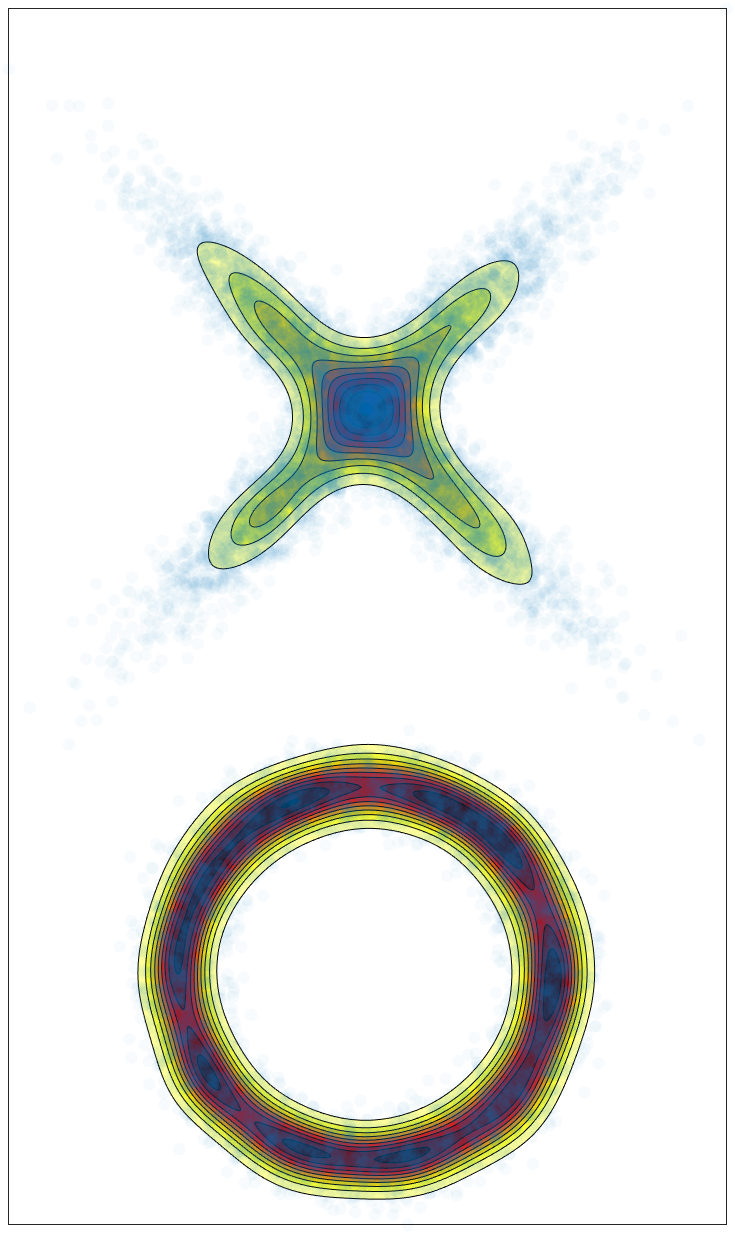}}\quad
    \subfloat{\includegraphics[width=0.195\textwidth]{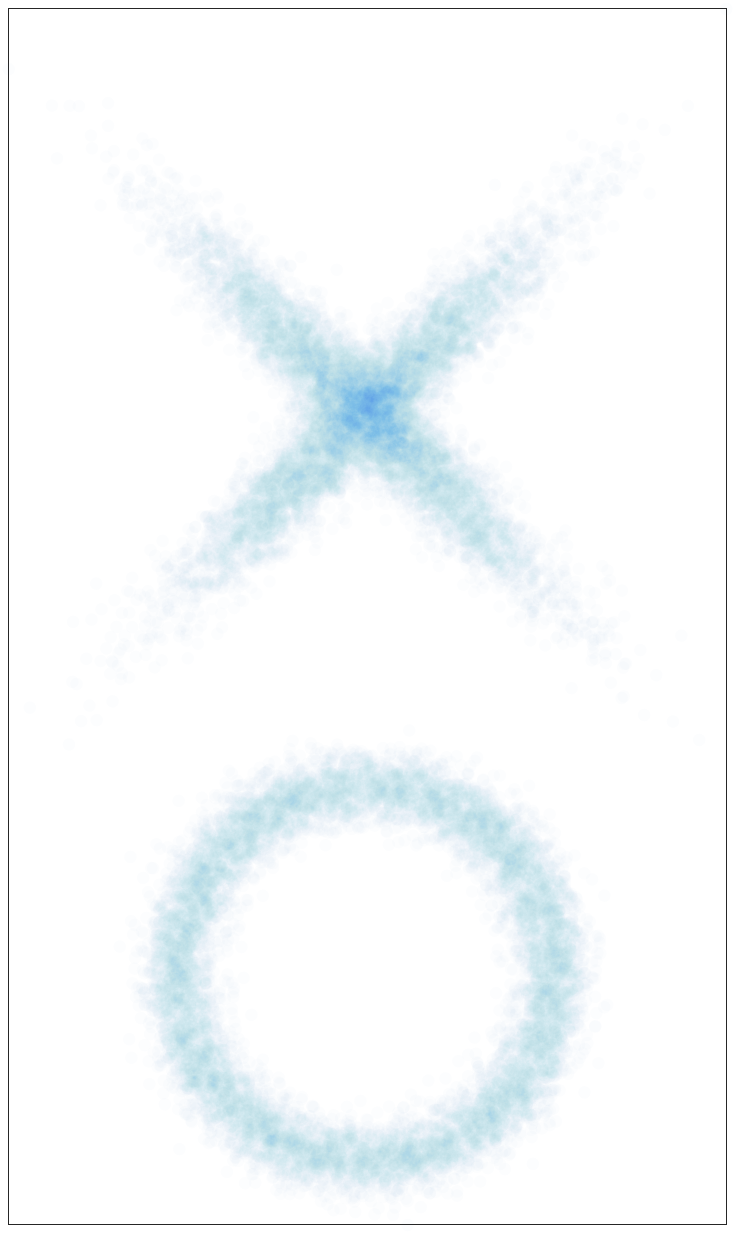}}\\\vspace{-10pt}
    \subfloat[MAP (Algorithm 1)]{\includegraphics[width=0.195\textwidth]{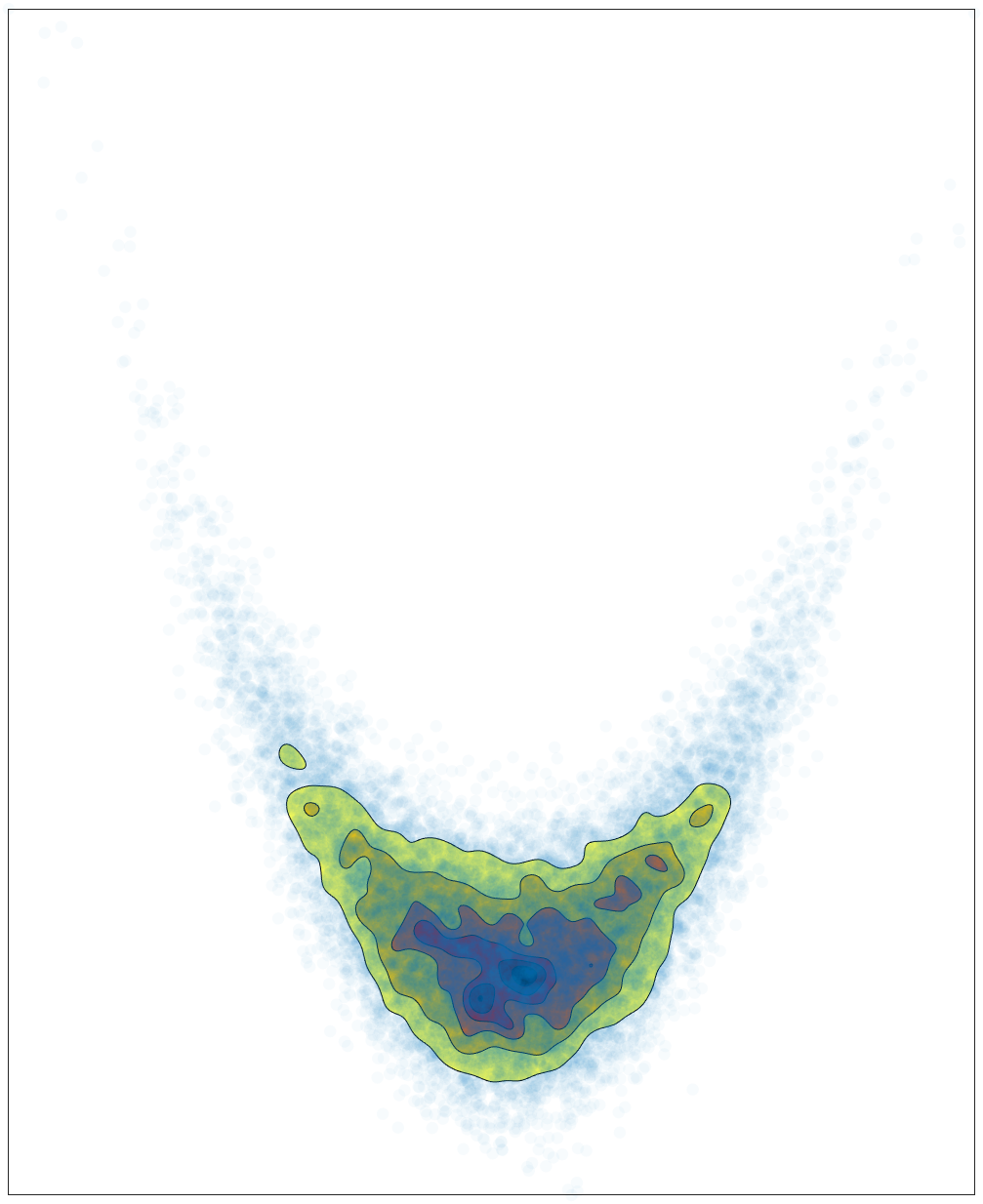}}
    \subfloat[FD (Algorithm 2)]{\includegraphics[width=0.195\textwidth]{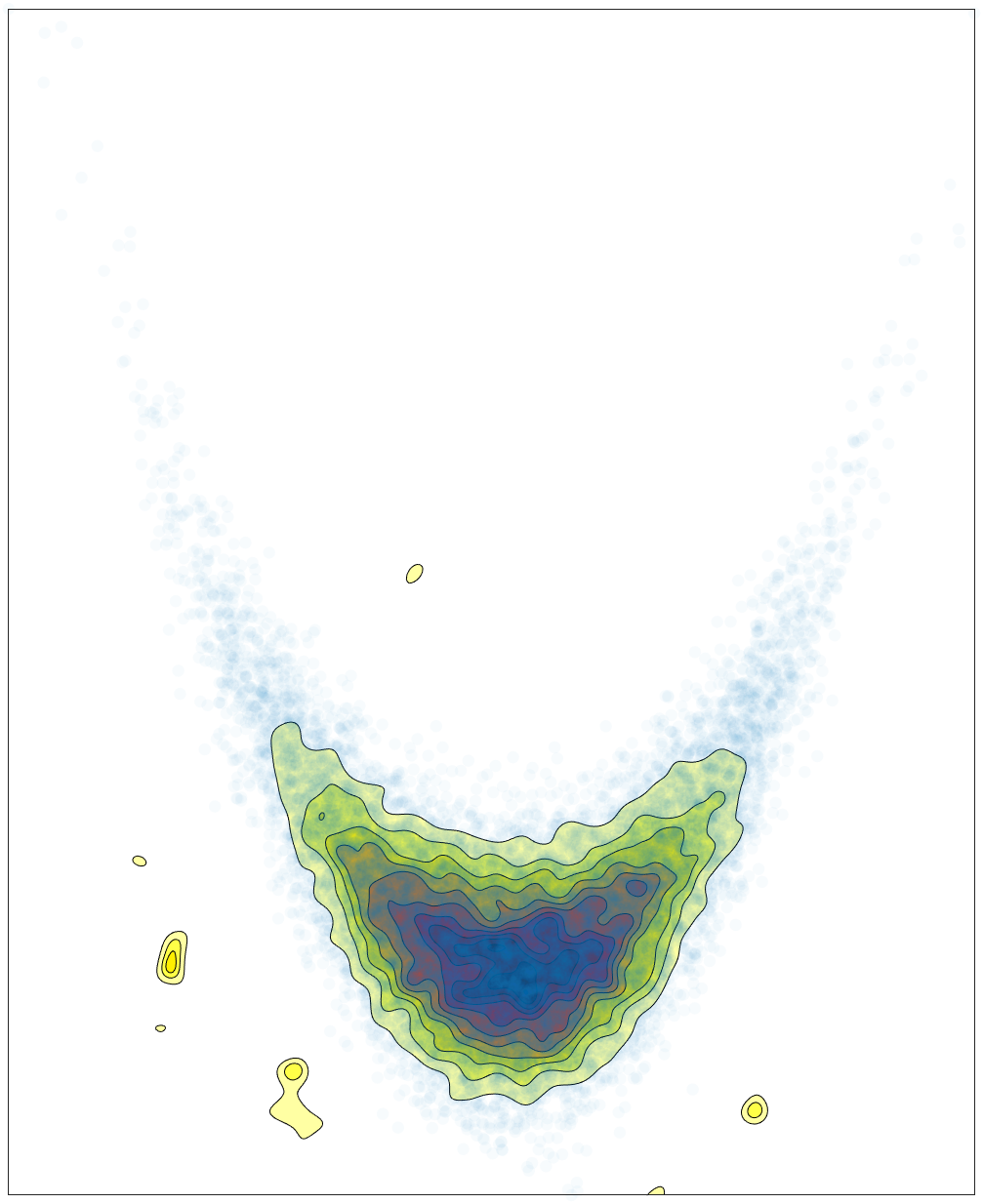}}
    \subfloat[NCFD (Algorithm 3)]{\includegraphics[width=0.195\textwidth]{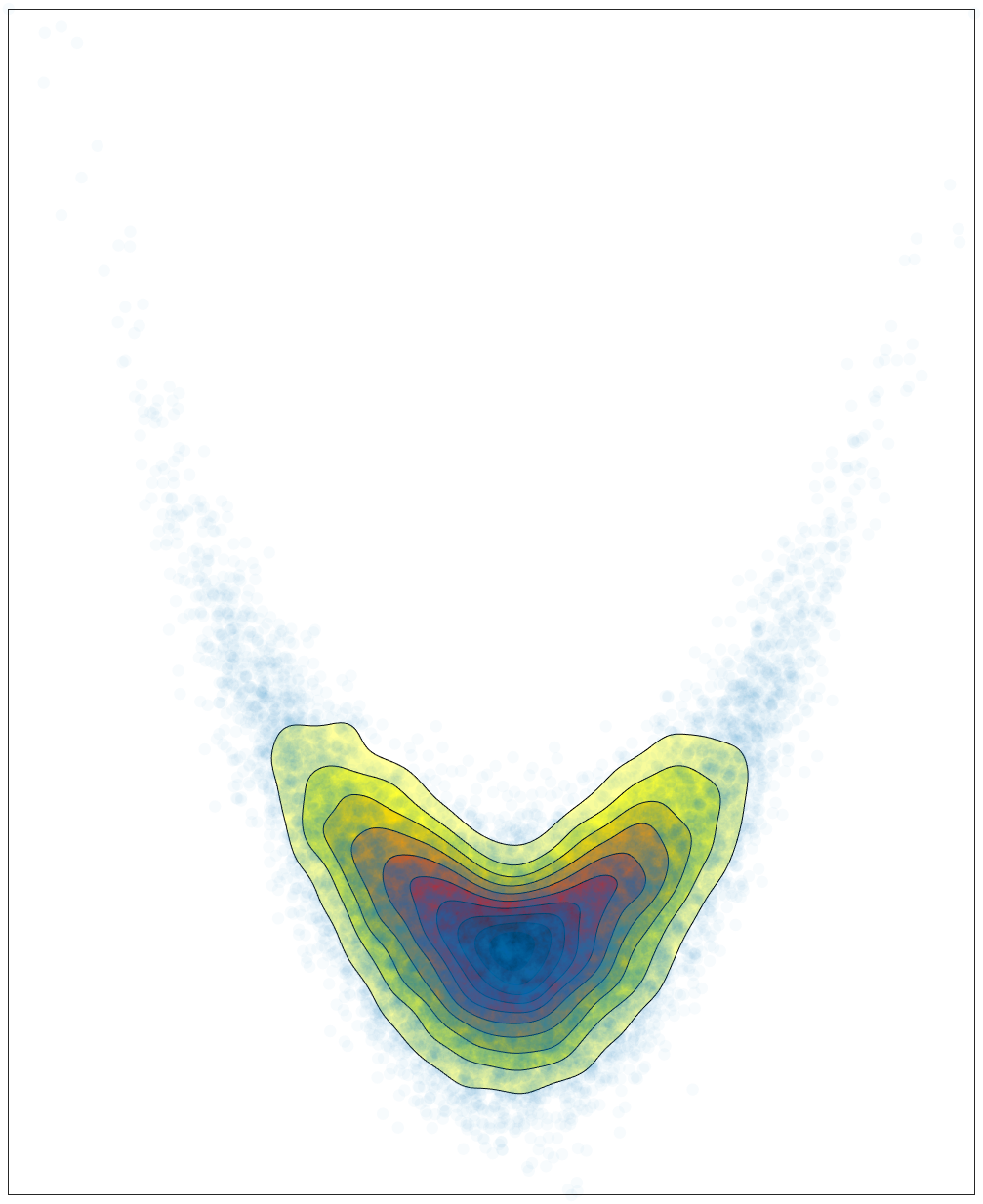}}
    \subfloat[FVPD (Algorithm 4)]{\includegraphics[width=0.195\textwidth]{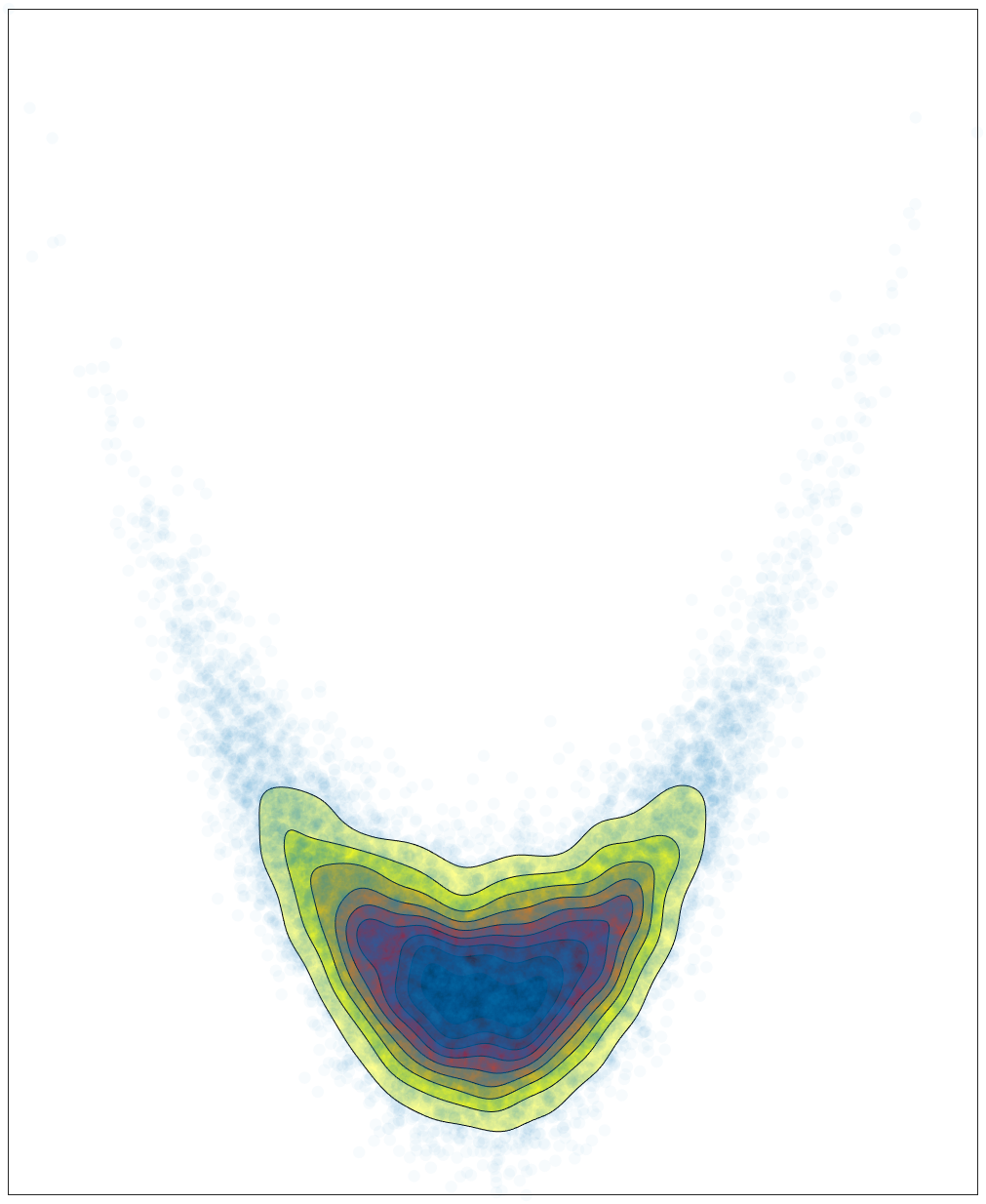}}\quad
        \subfloat[Data]{\includegraphics[width=0.195\textwidth,height=1.705in]{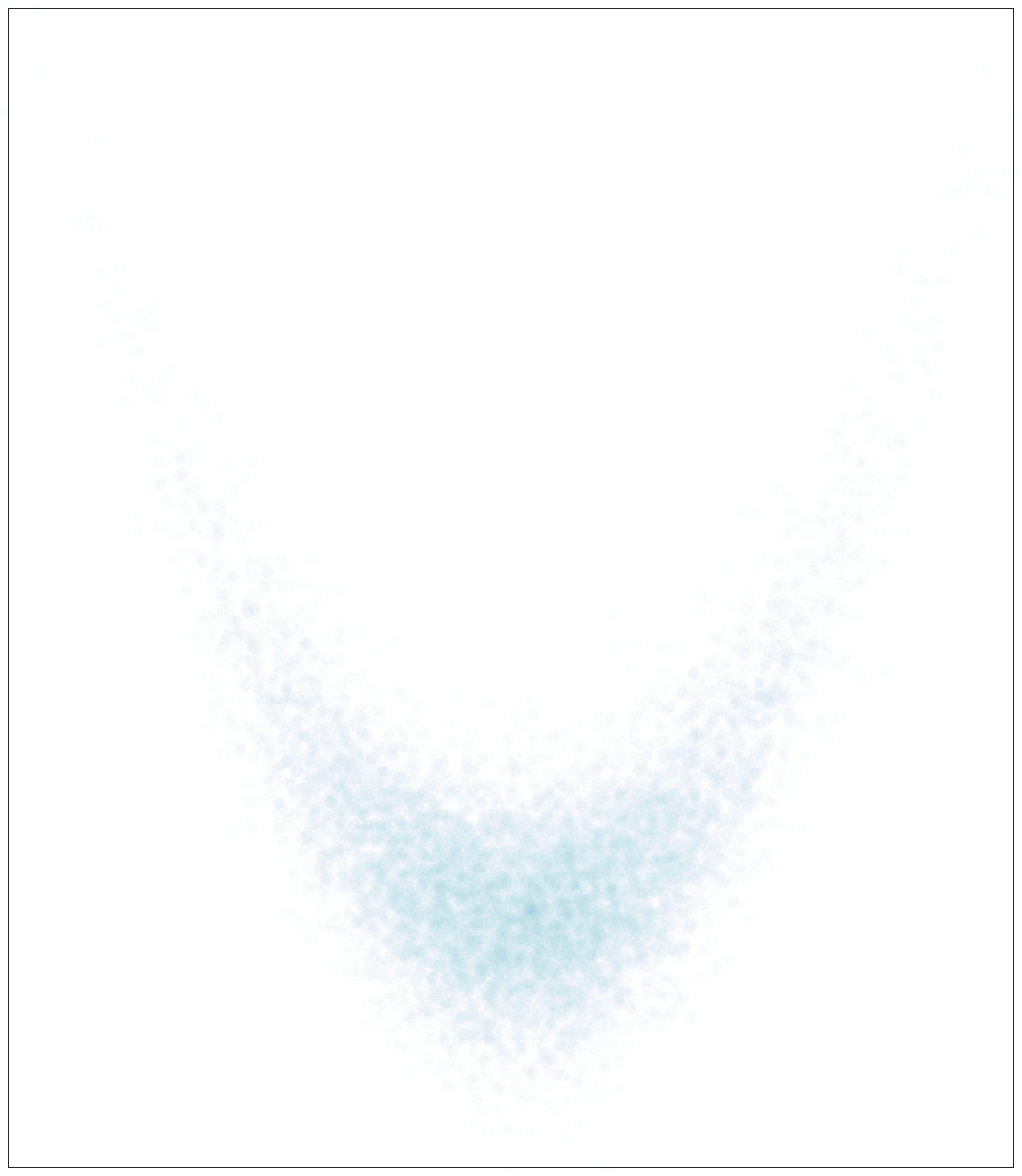}}
    \caption{Contour plots of GP-tilted density learned by each algorithm on three data sets.}
    \label{fig.toydata}
\end{figure*}

Since the data is $d$-dimensional, we compute these measures by first generating a direction $\boldsymbol{\mathrm{v}}$ uniformly distributed on the unit sphere in $\mathbb{R}^d$ and then calculate the marginal $\mathrm{KS}_{\boldsymbol{\mathrm{v}}}$, $\mathrm{WD}_{\boldsymbol{\mathrm{v}}}$ along the direction $\boldsymbol{\mathrm{v}}$ at $10^4$ grid points $\alpha\boldsymbol{\mathrm{v}}$ defined by scalars $\alpha$. We calculate the marginal distribution $\mathcal{P}_N$ by projecting the data on $\boldsymbol{\mathrm{v}}$, and calculate the marginal $Q$ along thin slices $\mathcal{B}_{\alpha\mathrm{v}}$ perpendicular to $\mathrm{v}$ around grid scalings $\alpha$ by approximating the unnormalized marginal $\int \boldsymbol{1}(x\in \mathcal{B}_{\alpha\mathrm{v}})f_{\mathrm{alg}}(x)\mathcal{N}(x|\mu,\Sigma) dx$ using 250K samples from the Gaussian base. Here, $f_{\mathrm{alg}}(x)$ is the exponential term in either (\ref{eq.density}) or (\ref{eq.predictive}), depending on the algorithm used. Sampling creates a discrete distribution proportional to $f_{\mathrm{alg}}(\hat{x}_s)\delta_{\hat{x}_s}$ at sample point $\hat{x}_s$. The \textit{weighted} marginal CDF can be found by projecting the $\hat{x}_s$ on $\boldsymbol{\mathrm{v}}$ and weighting by $f_{\mathrm{alg}}(\hat{x}_s)$. We use 500 samples of $\boldsymbol{\mathrm{v}}$ and center the data for these experiments.

In Table \ref{tab.quantitative}, we see that MAP often performs worse than at least one FD-based learning method. Also, noise-conditional and variational methods often improve upon the base FD algorithm. While NCFD and FVPD are competitive with each other, FVPD is roughly $H$ times faster than NCFD ($H=10$ here; see Table \ref{table:datasets}). In Figure \ref{fig.KSexamples} we show 10 example CDFs of random projections for MAGIC data using FVPD. 

In Table \ref{table:datasets} we show total run times for each algorithm. MAP is calculated over 10K iterations and is primarily a function of MC sample size rather than data size or dimension. We sample 100K points once and use the same samples for each iteration, significantly speeding up the MAP algorithm. FD and FVPD are both faster than NCFD roughly by a factor of parameter $H$, and since FVPD uses statistics calculated for FD with almost no extra overhead, they have the same run time. We can therefore see that FVPD has a significant time advantage as the data size grows.

\paragraph{Qualitative Results} We also illustrate TGP estimation on two dimensional data. First, we focus on dimensions 7 and 8 from the CA House data set due to it's unique pattern in a real world data set. In Figure \ref{fig:CA_samps} we show the original data with 10K samples generated from each learned density superimposed. In general, we see that the GP-tilted model is able to learn complex densities with a single vector $\theta\in\mathbb{R}^S$ where $S=1000$. MAP and FD qualitatively underperform NCFD and FVPD in opposite directions; FD overfits, while MAP underfits. Compared with the standard NCFD model, our proposed FVPD is able to learn sharper densities. We conjecture it may achieve this because it models ``noise'' (uncertainty) in the posterior of $\theta$ rather than in the noise-added data itself. As baselines, we show classic kernel density estimation (KDE) on this data, for both exact and RFF approximation, according to
$$ p(x|X) ~\propto~ \frac{1}{N}\sum\nolimits_{i=1}^N k(x,x_i) ~\approx~ \phi(x)^\top\Big[\frac{1}{N}\sum\nolimits_{i=1}^N \phi(x_i)\Big].$$ 
KDE-RFF performed poorly for smaller $S$ due to artifacts, which are still visible when $S=100,000$. The exact KDE also does well, but is based directly on the raw data and is computationally less appealing than a parameterized representation. In Figure \ref{fig:NCFD_CA} we also show the noise conditional density contours for (NCFD) as a function of decreasing $\sigma$, where we can see the multiscale advantage of that learning approach not available to FVPD.

In Figure \ref{fig.toydata} we show contour plots for our TGP learning algorithms using three other toy data sets: MNIST-60K projected onto its first two principal components, and XO and Smile synthetic data. For NCFD, we show the final data-level density for $\sigma=0$.  Noticeable in the FD algorithm is a tendency to learn high density values outside of the data region. This issue is addressed as intended by the other Fisher approaches, although we observe that NCFD has issues with XO. In these low dimensional toy data sets, we see that TGP is able to learn fairly complex densities without requiring a similarly complex density model. We also refer to Figure \ref{fig.KSexamples} for qualitative examples of how the learned densities in higher dimensions match the data by comparing their CDFs over 1D marginals.

\section{Conclusion}
We proposed a GP-tilted density for low dimensional estimation problems. Our Gaussian process makes use of the random Fourier feature representation for tractable nonparametric learning. We derived three Fisher divergence based learning algorithms, for which the GP-RFF score function is mathematically equivalent to a single layer neural network with known hidden layer parameters. Two features and contributions of our methods include:
\begin{enumerate}[leftmargin=*]
    \item With the exception of normalizing constants, all integrals and expectations are closed form, or easily approximated in closed form (Equation \ref{eqn.mgfapprox}). The algorithms are therefore not iterative, but use sufficient statistics of the data.
    \item We propose a variational approximation to the model posterior using an ELBO-like function based on the Fisher divergence and variational tempering. The corresponding predictive distribution on the data integrates out uncertainty in the model parameter rather than by adding noise to the data, giving an alternative to the noise conditional Fisher divergence for our model.
\end{enumerate}
Potential future developments include adapting and applying TGP within larger AI systems for real world applications and investigating methods for learning the important regularization parameters $\lambda$ (GP prior) and $\eta$ (variational tempering), and the kernel width $\gamma$. While we provided suggestions for setting $\eta$ as a function of $\gamma$, parameter settings were still empirically hand tuned and can be further optimized. Other neural network based density estimators may also benefit from incorporating the proposed GP learning framework \cite{zhang2019random,xu2024sparse}.

\bibliographystyle{IEEEtran}

\bibliography{GPdens}

\end{document}